\documentclass{article}

\usepackage{PRIMEarxiv}
\usepackage[utf8]{inputenc} 
\usepackage[T1]{fontenc}    
\usepackage{hyperref}       
\usepackage{url}            
\usepackage{booktabs}       
\usepackage{amsfonts}       
\usepackage{nicefrac}       
\usepackage{microtype}      
\usepackage{xcolor}         
\usepackage{amsmath}
\usepackage{algorithm}
\usepackage{algorithmic}
\usepackage{graphicx}
\usepackage{wrapfig}
\usepackage{caption}
\usepackage{subcaption}
\usepackage{graphics}

\newcommand{\IL}{I_L}
\newcommand{\ILi}{I_{L_i}}
\newcommand{\dotILi}

\title{Adaptive Federated Learning via Dynamical System Model 
\thanks{\textit{\underline{Citation}}: 
\textbf{Aayushya Agarwal, , Larry Pileggi, Gauri Joshi}} 
}

\author{
  Aayushya Agarwal, Larry Pileggi, Gauri Joshi \\
  Department of Electrical and Computer Engineering \\
  Carnegie Mellon University \\
  Pittsburgh, PA\\
  \texttt{\{aayushya, pileggi,gaurij\}@andrew.cmu.edu} \\
}

\begin{document}
\maketitle

\begin{abstract}
Hyperparameter selection is critical for stable and efficient convergence of heterogeneous federated learning, where clients differ in computational capabilities, and data distributions are non-IID. Tuning hyperparameters is a manual and computationally expensive process as the hyperparameter space grows combinatorially with the number of clients. To address this, we introduce an end-to-end adaptive federated learning method in which both clients and central agents adaptively select their local learning rates and momentum parameters. 
Our approach models federated learning as a dynamical system, allowing us to draw on principles from numerical simulation and physical design. Through this perspective, selecting momentum parameters equates to critically damping the system for fast, stable convergence, while learning rates for clients and central servers are adaptively selected to satisfy accuracy properties from numerical simulation. The result is an adaptive, momentum-based federated learning algorithm in which the learning rates for clients and servers are dynamically adjusted and controlled by a single, global hyperparameter. By designing a fully integrated solution for both adaptive client updates and central agent aggregation, our method is capable of handling key challenges of heterogeneous federated learning, including objective inconsistency and client drift. Importantly, our approach achieves fast convergence while being insensitive to the choice of the global hyperparameter, making it well-suited for rapid prototyping and scalable deployment. Compared to state-of-the-art adaptive methods, our framework is shown to deliver superior convergence for heterogeneous federated learning while eliminating the need for hyperparameter tuning both client and server updates
\end{abstract}


\section{Introduction}


Federated learning collaboratively trains a global model across decentralized clients without sharing raw data. In the presence of intermittent client availability, heterogeneous computational capabilities, and non-IID data distributions, federated learning can suffer from issues such as objective inconsistency \cite{fednova} and client drift \cite{shi2022optimization}, leading to degraded performance. As data and client heterogeneity increases, balancing efficient and stable global convergence across clients remains a challenge.


Prior work addresses these issues through aggregation adjustments \cite{feddyn, pathak2020fedsplit, karimireddy2020scaffold, li2020federated, li2019convergence, malinovsky2023server, charles2020outsized}, momentum-based updates \cite{das2022faster, xu2022coordinating, khanduri2021stem}, and Newton-like methods \cite{li2019feddane}. However, convergence of these client and server-side update strategies often depends on carefully tuned hyperparameters. The performance of federated learning is sensitive to the choice of hyperparameters such as learning rates and regularization terms \cite{kim2023adaptive, feddyn, li2019convergence}. The challenge of selecting appropriate hyperparameters is compounded as the hyperparameter search space scales combinatorially with the number of clients. Practical implementations often apply a uniform hyperparameter configuration across all clients to avoid searching the entire hyperparameter space; however, this approach often yields suboptimal performance due to data and computational heterogeneity. To address this, adaptive methods have been proposed, typically focusing on either client step size selection or server-side aggregations.


Server-side adaptive methods adjust learning rates and aggregation strategies to account for system heterogeneity. For example, SCAFFOLD~\cite{karimireddy2020scaffold} uses control variates to reduce client drift, while FedYogi, FedAdaGrad, and FedAdam~\cite{reddi2020adaptive} extend adaptive gradient methods to federated learning aggregation steps in order to stabilize noisy updates. FedExp~\cite{jhunjhunwala2023fedexp} applies exponential moving averages to smooth aggregation, and FedDyn~\cite{feddyn} introduces a dynamic regularizer to promote stability. However, these methods often rely on prior knowledge of client learning rates, limiting their ability to support client-specific and non-uniform step sizes. Moreover, these aggregation strategies introduce additional hyperparameters that affect the convergence and stability of the global model.

On the other hand, client-side adaptive methods focus on dynamically adjusting client updates to handle heterogeneity. FedProx~\cite{li2019convergence} adds a proximal term to limit client divergence, while MOON~\cite{li2021model} modifies client objectives to better align local heterogeneity. Auto-tuning methods like~\cite{kim2023adaptive} adapt client learning rates based on local gradients. Although these methods improve client optimization, without a coordinated server aggregation, these client-side methods alone can be prone to issues in heterogeneous federated learning including client drift and degraded performance.

In this work, we introduce a fully adaptive federated learning method that dynamically tunes hyperparameters for \emph{both} individual client updates and central server aggregation for efficient and stable convergence in heterogeneous settings. Our approach is based on a dynamical system model of federated learning \cite{fedecado} and draws on principles from numerical simulation and circuit design to select learning rates as well as momentum parameters. Through this perspective, constructing updates for heterogeneous federated learning is effectively translated into designing an adaptive simulation engine. Specifically, by mapping the dynamical system representation to an equivalent circuit, we are able to adopt well-established circuit design and simulation principles to choose step-sizes and momentum terms.

First, our methodology selects client learning rates based on numerical integration principles to ensure accuracy. This allows each client to operate with independent, data-specific learning rates that adapt to the local gradient space. To address objective inconsistency, we employ an integration/extrapolation operator at the central server's aggregation step that aligns the client updates in synchronous timescale. Then, on the server side, we adapt learning rate and momentum parameters using insights from circuit design, to promote fast and stable convergence.

\textbf{Our main contribution} is a fully adaptive, end-to-end federated learning method in heterogeneous settings where client and server updates are dynamically controlled by a single, global parameter. Importantly, the performance of our method is insensitive to this parameter’s value, removing the need for hyperparameter tuning and making the system practical to deploy at scale. We benchmark our approach across a wide range of federated learning scenarios with heterogeneity and non-IID data distributions. 
Our results show that our method achieves strong model performance without requiring hyperparameter tuning and outperforms prior adaptive methods that rely on carefully selected hyperparameters. This highlights the potential of our method in heterogeneous settings.

\section{Dynamical Model of Federated Learning}

The challenges associated with heterogeneous federated learning (such as client drift and objective inconsistency) and hyperparameter tuning stem from discrete client updates that struggle to handle variability in data and client computation. Rather than addressing these issues directly by tuning the hyperparameters of the discrete update, we take a different approach by modeling federated learning as a continuous-time dynamical system. 

The dynamical system model captures the evolution of both client and server states over continuous-time, with the goal of modeling the trajectory of the following global optimization problem:
\begin{equation}
    \min_x \sum_i p_i f_i(x, \mathcal{D}_i),
    \label{eq:obj_function}
\end{equation}
where $x$ represents the global model parameters and $f_i$ represents the local loss function of a client, $i$, characterized by a local dataset $\mathcal{D}_i$. For simplicity, we denote $f_i(x)\equiv f_i(x, \mathcal{D}_i)$. The local objective functions are weighted by a scalar, $p_i$, representing the size of their local dataset \cite{fedecado, fednova}.

We adopt a dynamical system model for solving \eqref{eq:obj_function} from \cite{fedecado} that represents the states of each client as $x_i(t)$ and the central agent model states as $x_c(t)$. To couple the client states with the central agent states, \cite{fedecado} introduces a coupling flow vector, $\ILi(t)$, that captures the \emph{accumulated} difference between the local client state $x_i$ and the central state $x_c$ over the simulation time according to:
\begin{equation}
\ILi(t) = L_i^{-1} \int_0^t \left( x_c(s) - x_i(s) \right) ds,
\label{eq:IL_integral_form}
\end{equation}
with its time derivative given by:
\begin{equation}
\dot{\ILi}(t) = L_i^{-1} \left( x_c(t) - x_i(t) \right).
\end{equation}

The coupling vector, $\ILi(t)$, is then integrated into the dynamical system modeling the continuous-time evolution of the client and central states:
\begin{equation}
        \frac{d}{dt}x_c(t) + \sum_{i=1}^{|\mathcal{C}|}\ILi(t) = 0 
    \label{eq:central_agent_ode}
\end{equation}
\begin{equation}
     L_i \dot{\ILi}(t) = x_c(t) - x_i(t) \quad \forall i\in[1,|\mathcal{C}|]
    \label{eq:inductor_ode}
\end{equation}
\begin{equation}
       -\ILi(t) + \frac{d}{dt}x_i(t) + p_i \nabla f_i(x_i(t)) =0 \quad \forall i\in[1,|\mathcal{C}|]
    \label{eq:client_ode}
\end{equation}
where $\mathcal{C}$ is the set of clients.

 $\ILi$ introduces a second-order dynamic that expedites the speed of convergence to steady-state \cite{ecado}. The hyperparameter $L_i$ represents a client-specific momentum term that controls how quickly the flow variable $\ILi(t)$ responds to discrepancies between the global and local states; larger values of $L_i$ slow the rate of change, introducing smoother, more stable dynamics.
 

The dynamical system naturally converges toward a steady-state that achieves global consensus across all clients even when their data and compute capacities differ. This naturally accounts for heterogeneity in clients.
When the system reaches equilibrium, the vector $\ILi \rightarrow0$, signifying that $x_i \rightarrow x_c$ for all clients. The rate at which the system settles is determined by the hyperparameter $L_i$, which can be tuned to accelerate convergence.

This dynamical system model is inspired by an analog circuit shown in Appendix \ref{app:circuit_background}, where $\IL$ is mapped to an electronic component known as an inductor with an inductance of $L_i$.

\subsection{Simulating the Dynamical System in Federated Learning Setting}

From the perspective of the dynamical system model, training a federated model in a heterogeneous environment becomes a matter of accurately simulating \eqref{eq:central_agent_ode}-\eqref{eq:client_ode} to its steady-state. Due to the distributed nature of federated learning, the simulation is partitioned across clients, where each computational node independently models the evolution of a client's state-variables and a central server aggregates these trajectories.

To distribute the simulation, we decouple the full set of coupled ordinary-differential equations (ODEs) \eqref{eq:central_agent_ode}-\eqref{eq:client_ode} using an iterative Gauss-Seidel (G-S). This method separates each client’s subproblem from the central agent by treating the coupling vector, $\IL^i$, as fixed from the previous iteration. 

At the $k+1$ G-S iteration, active clients, denoted by the set $\mathcal{C}_a$, solve their local ODEs by simulating:
\begin{equation}
\dot{x}_i^{k+1}(t) + p_i \nabla f_i(x_i^{k+1}(t)) + \ILi^{k}(t) = 0,
\label{eq:client_ode_gs}
\end{equation}
where $\ILi^{k}(t)$ is treated as a constant coupling term from the previous iteration. Each client is simulated for a client-specific time-window $[0,T_i]$ which varies based on local computational capabilities.

Each active client then communicates the final state $x_i^{k+1}(T_i)$ to the central agent, which then updates the global state, $x_c^{k+1},(t)$ and the coupling vector, $\IL^{i^{k+1}}(t)$, using the following coupled ODEs:
\begin{equation}
\dot{x}_c^{k+1}(t) = \sum_{i=1}^n \ILi^{k+1}(t),
    \label{eq:central_agent_capacitor_gs}
\end{equation}
\begin{equation}
    L_i \dot{\ILi}^{{k+1}}(t) = x_c^{k+1} - \ILi^{{k+1}}(t)\hat{G}_i^{th^{-1}} -x_i^{k+1}(t) +\ILi^{{k}}(t)\hat{G}_i^{th^{-1}}. 
    \label{eq:central_agent_inductor_gs}
\end{equation}

In this update, each client’s state $x_i^{k+1}$ is assumed constant over the server’s simulation.

The matrix $\hat{G}_i^{th}$ represents the \textit{first-order sensitivity} of client $i$ to changes in the global state. Inspired by Thevenin impedances in circuit theory (which characterize how the current from a circuit component responds to changes in the voltage), $\hat{G}_i^{th}$ models how a client's state is expected to evolve in response to updates in the central agent state, $x_c$. This sensitivity allows the central agent to anticipate each client’s response, leading to improved convergence as shown in \cite{fedecado}. The sensitivity matrix is:
\begin{align}
G_{th}^i &= \frac{\partial \ILi}{\partial x_i} = \frac{1}{\Delta t} + p_i \nabla^2 f_i(x_i),
\label{eq:client_thevenin_impedance}
\end{align}
where $\Delta t$ is the client step-size. The full derivation of $G_{th}^i$ is provided in \cite{ecado}.

Computing the Hessian, $\nabla^2 f_i$, at every G-S iteration is computationally expensive. Instead, we approximate \eqref{eq:client_thevenin_impedance} with a \textit{constant aggregate sensitivity} $\hat{G}_{th}^i$, computed by averaging the Hessian across a representative subset of local datapoints. This results in a client-sensitivity model as follows:
\begin{align}
\hat{G}_{th}^i &= \frac{1}{\Delta t} + p_i \bar{H}^i,
\label{eq:constant_sensitivity_def}
\end{align}
with $\bar{H}^i$ denoting the average Hessian over the sampled datapoints. 
In federated learning with non-IID data, each client’s loss is shaped by its unique data distribution, leading to distinct local objectives. The Hessian $\nabla^2 f_i(x_i)$ captures the curvature of each client’s loss and as a result, the sensitivity matrix, $\hat{G}_i^{th}$, directly reflects client data heterogeneity since non-IID distributions produce distinct sensitivity matrices across clients. However, computing the full Hessian for large-scale models can become a significant computational bottleneck. To address this, our approach is also compatible with efficient Hessian approximations, such as Fisher information matrices \cite{jhunjhunwala2024fedfisher}, as discussed in Appendix \ref{app:hessian}. Since the Hessian primarily serves as a weighting scheme across individual clients, these approximations are shown to have minimal impact on performance.

Using the continuous-time dynamical system model of federated learning, our goal is now to simulate the ODE for each client \eqref{eq:client_ode_gs} and server aggregation \eqref{eq:central_agent_capacitor_gs}-\eqref{eq:central_agent_inductor_gs} without relying on hyperparameters to achieve stable and efficient convergence under heterogeneity.

\section{Designing Adaptive Simulation for Heterogeneous Federated Learning}

We propose an end-to-end adaptive federated learning method that addresses the challenge of hyperparameter tuning in heterogeneous environments. Our approach is grounded in a continuous-time dynamical systems model \eqref{eq:central_agent_ode}–\eqref{eq:client_ode}, where both client updates and server aggregation are modeled as coupled ODEs. Unlike prior work \cite{fedecado}, which required carefully tuned client- and server-specific hyperparameters, we introduce a fully adaptive, momentum-driven federated learning method that eliminates the need for hyperparameter tuning. Our method achieves robust performance across a range of non-IID data distributions and varying client compute capacities. Specifically, we introduce:

\begin{enumerate}
\item \textbf{Client-side adaptation}: Each client independently simulates its local continuous-time dynamics defined in \eqref{eq:client_ode_gs}, using non-uniform, adaptive learning rates that are shaped by local gradient space and follow accuracy principles from numerical simulation.
\item \textbf{Server-side adaptation}: The server updates its state using client-specific momentum terms, which are chosen to expedite global convergence. Then, the server-side aggregation is performed by a provably stable numerical method with adaptive step-sizes to handle non-uniform client learning rates (to avoid objective inconsistency).  
\end{enumerate}

This framework enables stable and efficient training under heterogeneous conditions, and achieves fast convergence without requiring hand-tuned parameters.

\subsection{Adaptive Client Updates}

By viewing the evolution of client states as a continuous-time model, our goal is to accurately simulate the client ODE. The client ODE in \eqref{eq:client_ode_gs} is simulated by marching forward in continuous time at discrete time-steps (or step-sizes) of $\Delta t_i$. The client state at the next time point, $t + \Delta t_i$, is defined by:
\begin{equation}
x_i(t+\Delta t) = x_i(t) + \int_{t}^{t+\Delta t_i} \ILi^k(\tau) + \nabla f_i(x_i(\tau), \mathcal{D}_i), d\tau.
\label{eq:client_ode_integral}
\end{equation}

In general, the integral in \eqref{eq:client_ode_integral} does not have a closed-form solution and instead is approximated using numerical integration methods. To minimize computational overhead, we opt for an explicit Forward-Euler (FE) integration method as follows:
\begin{equation}
x_i(t+\Delta t) = x_i(t) + \Delta t_i \left( \ILi^k(t) + \nabla f_i(x_i(t), \mathcal{D}_i) \right),
\label{eq:client_ode_fe}
\end{equation}
where $\Delta t_i$ is the client-specific time step. While the update resembles gradient descent with learning rate $\Delta t_i$, it is derived from numerical simulation. As a result, $\Delta t_i$ is not tuned for convergence to local minimum, but rather selected to ensure an accurate numerical approximation of the continuous-time dynamics in \eqref{eq:client_ode_integral}. This is guided by error bounds from numerical analysis.

The accuracy of a FE step is measured using the local truncation error (LTE), which captures the difference between the true ODE solution and its numerical approximation. The LTE for FE, denoted as $\varepsilon_{FE}$, is estimated as:
\begin{multline}
\varepsilon_{FE} = \frac{\Delta t_i}{2} \vert \ILi(t) + \nabla f_i(x_i(t), \mathcal{D}_i) - \ILi(t-\Delta t) -\\ \nabla f_i(x_i(t-\Delta t), \mathcal{D}_i) \vert.
\end{multline}
The derivation is provided in Appendix \ref{app:lte_fe_derivation}. To ensure the discretized steps using FE accurately follow the client ODE in \eqref{eq:client_ode_gs}, we aim to select $\Delta t_i$ to maintain the LTE below a pre-defined tolerance $\gamma$:
\begin{equation}
\max(\varepsilon_{FE}) \leq \gamma.
\label{eq:client_lte_condition}
\end{equation}

This follows standard numerical simulation practices, where step sizes are selected based on local error control. To ensure efficient simulation, we use a backtracking line search to find the largest $\Delta t_i$ satisfying \eqref{eq:client_lte_condition} (Appendix \ref{app:fe_adaptive_time_step}).
However, Forward-Euler (and similarly gradient descent) updates are prone to divergence as numerical errors accumulate over time \cite{cohen2024understanding}.
We adopt prior work from circuit simulation \cite{rohrer1981passivity} that determines the following $\Delta t$ that guarantees Forward-Euler stability:
\begin{equation}
\Delta t_{FE} \leq 2 \frac{x_i^{\top} (\nabla f_i(x_i(t))+\ILi)}{ \dot{x}_i^\top(t)\dot{x}_i(t)}.
\label{tfe-naive}
\end{equation}

 We use this bound as an initialization for adaptive step size selection, which is then refined through backtracking to improve LTE accuracy. This creates a client-specific adaptive step size that responds to the local geometry of each client’s loss function. 

In this work, we assume each active client performs local updates for a fixed wall-clock duration. Due to heterogeneity in compute capabilities, this leads to varying numbers of local epochs and learning rates. This corresponds to each client simulating its local ODE over a total time window of:\begin{equation}
T_i = \sum_{j=1}^{e_i} \Delta t_i,
\end{equation}
where $e_i$ is the number of local epochs determined by the client’s computational capacity. To accurately coordinate active client updates, clients report their simulation window $T_i$ and $x_i(T_i)$.

\subsection{Adaptive Server-Side Aggregation with Momentum}

At each communication round, the central server receives both the final simulated state, $x_i(T_i)$, and the simulation window, $T_i$, from each active client. These updates are aggregated according to the global ODE defined in \eqref{eq:central_agent_capacitor_gs}–\eqref{eq:central_agent_inductor_gs}. Our goal is to design a server-side aggregation that is insensitive to hyperparameters by developing an adaptive simulation that accurately tracks the central agent ODEs.

This is accomplished in two steps. First, we design the momentum parameters, $L_i$, to ensure fast convergence in continuous-time. Then, we develop an adaptive simulation engine that accurately follows the designed ODE for stable convergence to the steady-state.

\subsubsection{Selecting Momentum Parameters}
\label{sec:momentum_params}
The central agent dynamics, defined in equations \eqref{eq:central_agent_capacitor_gs}–\eqref{eq:central_agent_inductor_gs}, follow a linear ODE where the momentum parameter, $L_i$, plays a key role in both stability and convergence. This reflects a well-known challenge in momentum-based optimization (e.g., Nesterov’s acceleration), where performance is sensitive to the choice of momentum-related hyperparameters. From a dynamical systems perspective, $L_i$ functions as \emph{damping coefficients} that influences the convergence of the global state, $x_c(t)$, by shaping the eigenvalues of the system’s linearized dynamics.

\begin{figure}
\centering
    \includegraphics[width=0.7\linewidth]{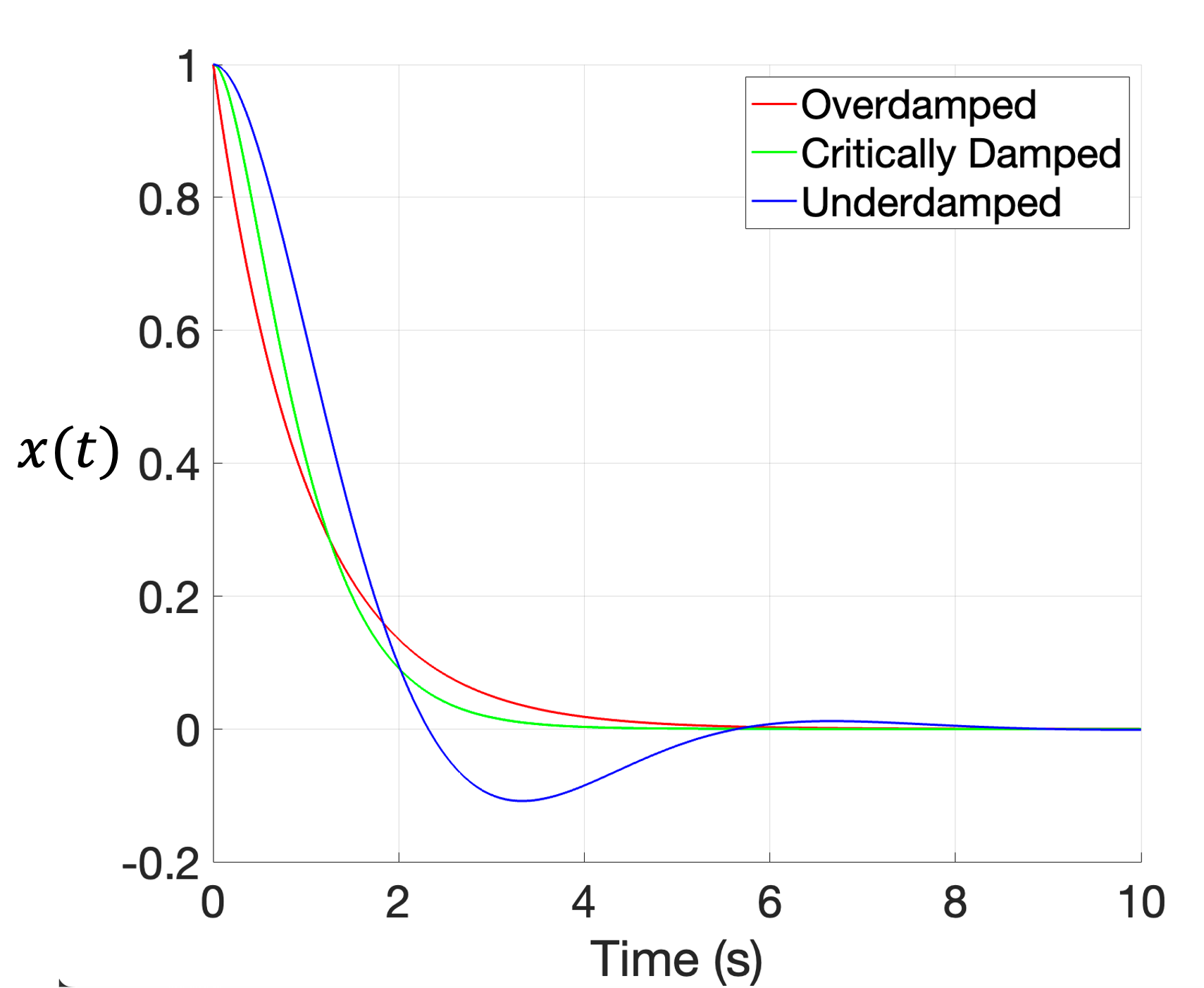}
    \caption{\small Simulating a single-variable second-order ODE, $\ddot{x}(t)+L\dot{x}+10x(t)=0$, we vary the parameter $L$ to achieve overdamped, critically damped, and underdamped systems.}
    \label{fig:simple_rlc_damping}
\end{figure}

These systems can be designed to fall into one of three damping regimes as illustrated in Figure \ref{fig:simple_rlc_damping}:
\begin{enumerate}
    \item \textbf{Over-damped}: The system converges without oscillation and overshooting but more slowly, as the damping suppresses responsiveness. 
    \item \textbf{Under-damped}: The system converges with oscillatory behavior.
    \item \textbf{Critically damped}:  The system converges quickly to steady-state without oscillation.
\end{enumerate}

Our goal is to design the central agent ODE to be \emph{critically damped} for fast convergence to steady-state without oscillations. Optimally selecting $L_i$ to achieve this requires controlling the eigenvalues of the system to lie at the boundary between overdamping and underdamping.

Mathematically, this can be achieved by optimizing:
\begin{align}
\label{eq:critical_damping_eig}
    \max_L inf \left( Re(-eig(\begin{bmatrix}
        C, 0\\0, L
    \end{bmatrix}^{-1} G )) \right) \\
    s.t. \quad Im(-eig(\begin{bmatrix}
        C, 0\\0, L
    \end{bmatrix}^{-1} G )) =0 
\end{align}

where $eig(\cdot)$ is the eigenvalue and $Re(\cdot)$ and $Im(\cdot)$ capture the real and imaginary components. $G$ is the collective client sensitivities. The full derivation is in Appendix \ref{app:max_eig_derivation}

Solving \eqref{eq:critical_damping_eig} is computationally challenging, as eigenvalue computation is impractical with a large number of clients. Rather, we develop an approximate solution that uses the structure of federated learning to select a momentum parameter, $L_i$, that puts the system close to critical damping. To develop an approximate solution, we map the dynamical system to an analog circuit in Appendix \ref{app:circuit_background} where the momentum parameter translates into selecting an inductance. This enables us to use tools from circuit theory to design the inductance to achieve critical damping.

Our approach is inspired by Thevenin impedances from circuits, which is used to simplify the behavior of complex interconnected components. By treating each client as a branch in a larger system connected through a central aggregator, we can apply similar techniques to analyze how each client’s flow vector, $\ILi$, (modeled by inductor current) respond to the shared central state $x_c$.

We study the sensitivity of the coupling variable, $\ILi$, with respect to the central state, $x_c$, as:
\begin{equation}
    \frac{\partial}{\partial x_c} \ILi(t) = -\frac{d}{dx_c}\dot{x}_c(t) - \sum_{j\neq i} \frac{\partial}{\partial x_c}I_{L_j}(t)
\end{equation}
This sensitivity is a combination of
\begin{enumerate}
    \item The influence of $\ILi$ on $\dot{x}_c$ (i.e., $\frac{d}{dx_c}\dot{x}_c(t)$) which is the direct path through which client $i$ influences the central state, and
    \item The indirect influence of $\ILi$ on all other clients’ inductor currents $\frac{\partial}{\partial x_c}I_{L_j}(t)$ ($j \neq i$), via their shared connection to $x_c$.
\end{enumerate}


For federated learning, the dominant contribution to this sensitivity comes from changes in the central agent, $\dot{x}_c$ has a significantly larger influence on $\ILi$ than the other clients (i.e., $\frac{d}{dx_c}\dot{x}_c(t) \gg \sum_{j\neq i} \frac{\partial}{\partial x_c}I_{L_j}(t)$). This is numerically demonstrated in Appendix \ref{app:critically_damped_L}.

This approximation is justified by how federated learning operates: clients do not communicate directly with one another but only through the central server. Updates from a client affect the global model, which then influences other clients in future rounds. Thus, at each step, the most immediate and significant effect of a client’s update is on the central agent, not other clients. This allows us to approximate the sensitivity of the system to $\IL^i$ using only the central agent’s response.

This approximation effectively decouples the clients from one another, allowing each to be analyzed independently in relation to the central server. Specifically, in equation \eqref{eq:central_agent_inductor_gs}, each client “sees” only the dynamics of the central agent’s capacitor, $\dot{x}_c(t)$, rather than interacting with other clients. As a result, we can model each client branch as an isolated second-order system as:
\begin{equation}
    \label{eq:simplified_client_branch}
\dot{x}_c^{k+1}(t) = \ILi^{k+1}(t),
\end{equation}
\begin{equation}
    L_i \dot{\ILi}^{{k+1}}(t) = x_c^{k+1} - \ILi^{{k+1}}(t)\hat{G}_i^{th^{-1}} -x_i^{k+1}(t) +\ILi^{{k}}(t)\hat{G}_i^{th^{-1}}. 
\end{equation}
which can be mapped to a series RLC circuit (Figure \ref{fig:series_rlc}) as derived in Appendix \ref{app:critically_damped_L}.

This decoupled view allows the momentum parameter $L_i$ to be designed independently for each client, without accounting for client interactions. By modeling each client-server connection in isolation, we can independently select $L_i$ to achieve critical damping. Using the RLC circuit model, we derive a closed-form expression for $L_i$ in Appendix \ref{app:critically_damped_L} that ensures critical damping:

\begin{equation}
L_i = \frac{1}{4} \hat{G}_{th}^{i^{-2}}.
\label{eq:critical_L}
\end{equation}

 The result is a momentum term that guarantees fast, stable convergence for the central agent dynamics.

\subsubsection{Adaptive Aggregation Updates}

With momentum parameters set in \eqref{eq:critical_L}, our next goal is to design an adaptive aggregation method for heterogeneous client updates. This involves building an adaptive simulation engine that (a) aligns client updates on a shared time scale to ensure objective consistency, and (b) uses a stable numerical integration scheme that adaptively selects the central agent’s step size to accurately follow the ODE.

\paragraph{Interpolating/Extrapolating Heterogeneous Updates to a Synchronous Timescale:} Due to client heterogeneity, active clients simulate their local ODEs for different simulation windows, $T_i$. Without coordinated aggregation, this mismatch can lead to objective inconsistency \cite{fednova}.  To align the client updates on a synchronous timescale, we uses a linear interpolation/extrapolation operator, $\Gamma(x_i(t), \tau)$, that evaluates each client’s state at intermediate timepoints $\tau$:
\begin{equation}
\Gamma(x_i(t),\tau) = \frac{x_i(t_2)-x_i(t_1)}{t_2 - t_1} (\tau - t_1)  + x_i(t_1),
\label{eq:interpolation_operator}
\end{equation}
This allows the central agent to evaluate its state over a synchronized window $\tau \in [t_0, t_0 + \max(T_i)]$, ensuring client updates are properly aligned. Without this, mismatched timescales would prevent convergence to a shared steady state, as shown in \cite{fedecado}. The central agent dynamics are then:
\begin{equation}
\label{eq:central_agent_capacitor_gs_interpolation}
\dot{x}_c^{k+1}(\tau) = \sum_{i=1}^n \ILi^{{k+1}}(\tau)
\end{equation}
\begin{multline}
L_i\dot{\ILi}^{{k+1}}(\tau) = x_c^{k+1}(\tau) - (\ILi^{{k+1}}(\tau)G_i^{th^{-1}} +\\  \Gamma(x_i^{k+1}(t),\tau) - \ILi^{{k}}G_i^{th^{-1}}).
\label{eq:central_agent_inductor_gs_interpolation}
\end{multline}

\paragraph{Simulating Central Agent via Backward-Euler:} Next, the central agent ODE \eqref{eq:central_agent_capacitor_gs_interpolation}-\eqref{eq:central_agent_inductor_gs_interpolation} is numerically simulated to solve for the central agent states. We propose using a numerically stable, Backward-Euler method that defines each aggregation step as:
\begin{equation}
x_c^{k+1}(\tau+\Delta t) = x_c^{k+1}(\tau) - \Delta t \sum_{i=1}^{n} \ILi^{{k+1}}(\tau+\Delta t),
\label{eq:central_agent_be_cap}
\end{equation}
\begin{multline}
\ILi^{{k+1}}(\tau+\Delta t) = \ILi^{{k+1}}(\tau) + \frac{\Delta t}{L} \Big( x_c^{k+1}(\tau+\Delta t) - \\( \ILi^{{k+1}}(\tau+\Delta t) G_i^{th^{-1}}
+ \Gamma(x_{i}^{k+1}(t),\tau+\Delta t) - \\\ILi^{k}(\tau+\Delta t) G_i^{th^{-1}} ) \Big).
\label{eq:central_agent_be_ind}
\end{multline}

The Backward-Euler method is unconditionally stable as derived in Appendix \ref{app:be_derivation}, meaning it converges to the system’s steady-state (i.e., critical point of the objective) regardless of the step size $\Delta t$. 
\paragraph{Adaptive Time-Step Selection} Although the BE provides numerical robustness, our objective is to select a $\Delta t$ that accurately follows the central agent ODE designed for efficient convergence.
We introduce an adaptive time-stepping scheme for the central agent that selects the step size, $\Delta t$, based on the numerical accuracy of the BE rather than treating it as a hyperparameter.

The accuracy of the BE update is measured using the local truncation error (LTE). For the central agent dynamics in \eqref{eq:central_agent_capacitor_gs_interpolation}, the LTE is estimated as:
\begin{equation}
\varepsilon_{BE}^C = -\frac{\Delta t}{2}\left[ \sum_{i=1}^{n} \ILi^{{k+1}}(\tau) - \sum_{i=1}^{n} \ILi^{{k+1}}(\tau+\Delta t) \right].
\label{eq:be_lte_cap}
\end{equation}

The LTE for a BE integration of \eqref{eq:central_agent_inductor_gs_interpolation} is:
\begin{multline}
\varepsilon_{BE_i}^L = -\frac{\Delta t}{2L} \Big[ (x_c^{k+1}(t) - \ILi^{{k+1}}(t)\bar{G}_i^{-1} + x_i^{k+1}(t) - \\ \ILi^{k}(t)\bar{G}_i^{-1}) -
(x_c^{k+1}(t+\Delta t) - \ILi^{{k+1}}(t+\Delta t)\bar{G}_i^{-1} +\\ x_i^{k+1}(t+\Delta t) - \ILi^{k}(t+\Delta t)\bar{G}_i^{-1}) \Big].
\label{eq:be_lte_ind}
\end{multline}

To ensure accurate simulation, we adaptively choose $\Delta t$ using a backtracking line search (Appendix~\ref{app:be_adaptive_time_step}) such that the maximum LTE across all components remains below a pre-defined tolerance $\gamma$:
\begin{equation}
\max |\varepsilon_{BE}| \leq \gamma, \quad \varepsilon_{BE} = [\varepsilon_{BE}^C, \varepsilon_{BE}^L].
\end{equation}

\subsection{Adaptive End-to-End Federated Learning}

Our method is an adaptive, end-to-end federated learning algorithm designed to address both non-IID data and heterogeneous client computation. Computational heterogeneity is captured through an interpolation/extrapolation operator that aligns client updates, while data heterogeneity is modeled through the aggregate sensitivity matrix, which encodes differences in local curvature induced by non-IID data distributions.

A key element of our approach is the circuit perspective. By framing the system as an equivalent circuit, we can (i) visualize the topology of client–server interactions, and (ii) directly apply circuit design principles to construct the momentum term such that the overall system is critically damped. This perspective also enables us to leverage circuit simulation techniques to select stable step sizes, ensuring the discretized system remains passive and adapts naturally to both client and server gradient spaces.

This construction allows step sizes across clients and the server to be governed by a single global hyperparameter, $\gamma$, which controls the accuracy tolerance for numerical integration. While $\gamma$ is tunable, we show that the algorithm remains numerically stable across its range. The robustness of our method stems directly from the use of Backward-Euler integration. Unlike explicit methods, Backward-Euler is unconditionally stable and guarantees convergence regardless of the step size. As proven in Appendix \ref{app:be_derivation}, this property ensures that our algorithm converges to a stable optimum while tracking the continuous-time ODE dynamics designed by the momentum parameter $L_i$. Robustness is further reinforced by monitoring the local truncation error (LTE), as detailed in Appendix \ref{app:be_adaptive_time_step}, which preserves the intended ODE behavior even under heterogeneous conditions.

The complete algorithm is provided in Appendix \ref{app:adaptive_fedecado_algo}.

\section{Experiments}

Our methodology, called Adaptive FedECADO, is an adaptive federated learning method that achieves fast convergence via critical damping, while remaining robust to hyperparameter selection in heterogeneous settings. We study the effectiveness of our approach under heterogeneity in both data and computation. In our experiments, client data distributions follow a non-IID Dirichlet distribution, $|\mathcal{D}i| \sim \text{Dir}{16}(0.1)$, and each client performs a random number of local training epochs sampled from a uniform distribution according to $e_i \sim U[1,50]$. 

In these settings, Adaptive FedECADO achieves high model performance without requiring any manual tuning. In contrast, prior adaptive methods including SCAFFOLD \cite{karimireddy2020scaffold}, FedAdam \cite{reddi2020adaptive}, FedAdaGrad \cite{reddi2020adaptive}, delta-SGD \cite{kim2023adaptive}, FedCAda \cite{zhou2025fedcada}, and FAFED \cite{wu2023faster} are sensitive to hyperparameter selections, leading to instability or poor convergence. Our results show that Adaptive FedECADO outperforms these baselines across multiple datasets and models, with faster convergence across a wide range of hyperparamter values.

\begin{table*}[h!]
    \centering
        \caption{ Percentage of useable models, defined as achieving over 80\% of the highest accuracy, across all hyperparameter selections for CIFAR-10, CIFAR-100, and Sentiment-140 datasets.}
    \label{tab:useable_models}
\resizebox{\columnwidth}{!}{%
    \begin{tabular}{|c|c|c|c|c|c|c|c|c|}
    \hline
    Dataset (Model) & Adaptive FedECADO & SCAFFOLD & FedAdam & FedAdaGrad & FedExp & deltaSGD & FedCAda & FAFED\\
    \hline
        CIFAR-10 (Resnet-18) & 100 & 42 & 15 & 15 & 0.05 & 50 & 75& 90 \\
        \hline
        CIFAR-100 (Resnet-50) & 100 & 15 & 0.05 & 0.05 & 0 & 30& 85& 70 \\ \hline
        Sentiment-140 (VGG-11) & 100 & 82 & 85 & 82 & 90 & 80 & 95& 85 \\
        \hline
             \end{tabular}%

}
\end{table*}

\subsection{Robustness to Hyperparameter Selection}

To evaluate robustness to hyperparameter selection, we perform a random search to identify the best-performing hyperparameters for each baseline method and for Adaptive FedECADO individually across multiple datasets and models, including CIFAR-10, CIFAR-100, and Sentiment-140 (shown in Appendix \ref{app:sentiment_140}). Figure \ref{fig:test} shows the final classification accuracies for each trial across the hyperparameters ranges reported in Appendix \ref{app:hyperparameter_ranges}.

Adaptive FedECADO is shown to consistently achieve high accuracy with minimal variance, demonstrating strong robustness to its single hyperparameter, $\gamma$. In contrast, baseline methods show high sensitivity, performing well only under carefully tuned settings. Nearly all Adaptive FedECADO runs reach near-optimal performance, effectively removing the need for hyperparameter tuning. This is a combined result of (a) selecting momentum parameters that accelerate convergence and (b) adaptive numerical methods that ensure stable convergence. The efficacy of each of these are shown in an ablation study in Appendix \ref{app:ablation}. While Adaptive FedECADO introduces a slight increase in computational cost due to the dynamical system formulation and Backward-Euler steps (as shown in Appendix \ref{app:runtime}), we argue that this overhead is justified by the method’s improved robustness across hyperparameter settings, particularly as extensive hyperparameter tuning itself can incur greater computational cost in practice.

\begin{figure}[h]
\centering
\begin{subfigure}{.45\textwidth}
  \centering
  \includegraphics[width=1\linewidth]{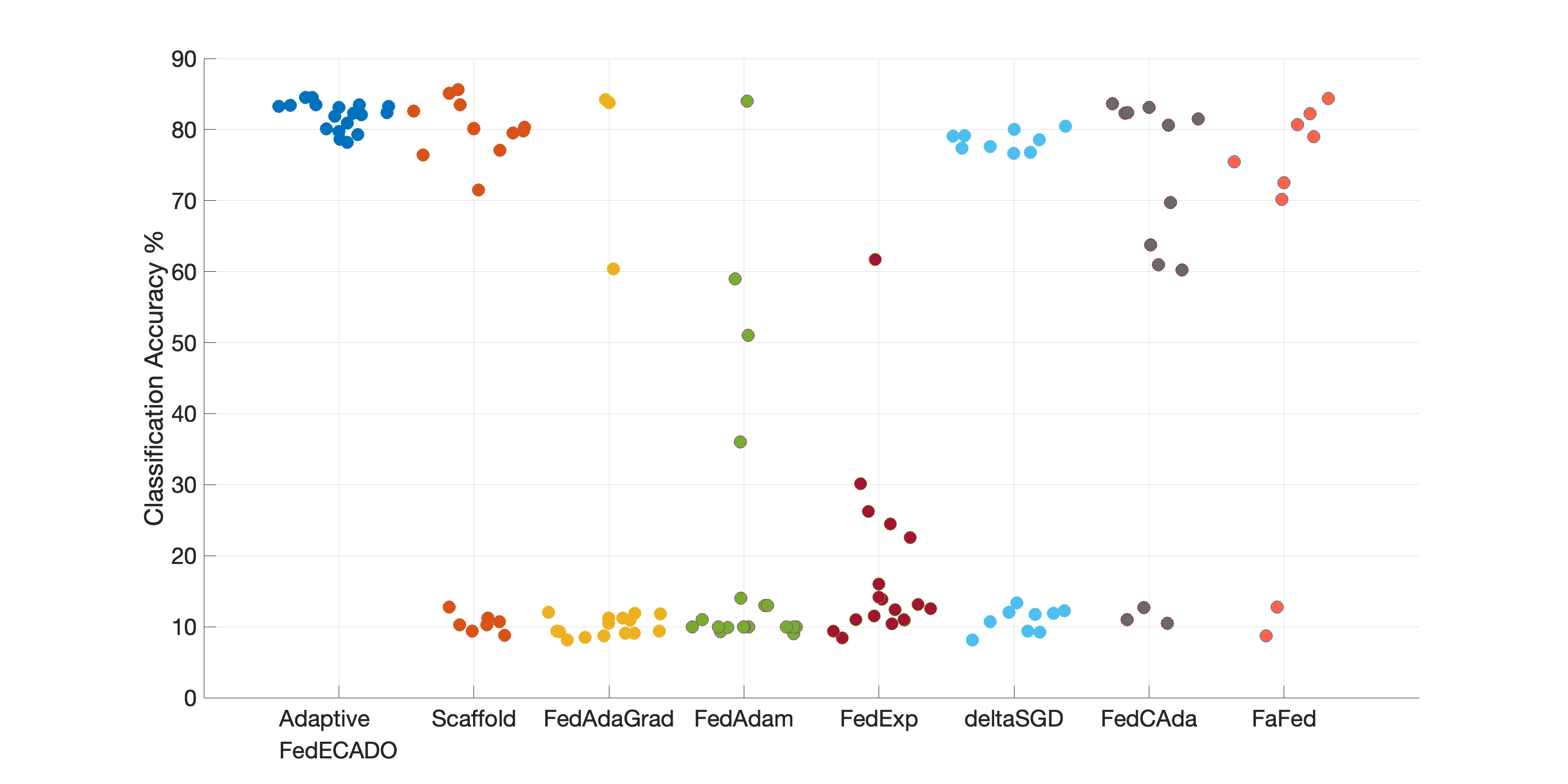}
  \label{fig:cifar10_sweep}
\end{subfigure}%
\hfill
\begin{subfigure}{.45\textwidth}
  \centering
  \includegraphics[width=1\linewidth]{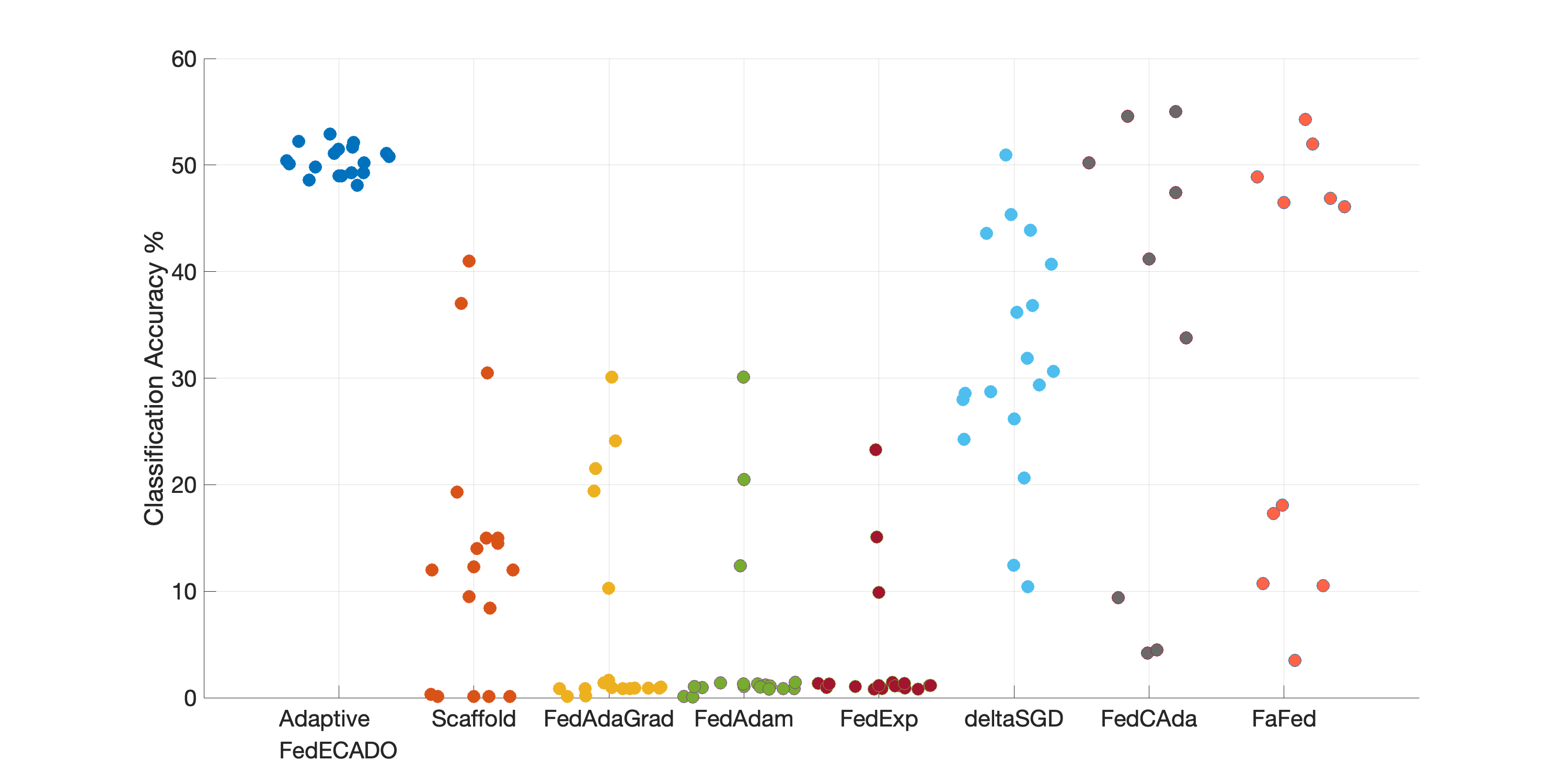}
  \label{fig:cifar100_sweep}
\end{subfigure}
\vspace{-2em}
\caption{Classification accuracy of Adaptive FedECADO compared to baseline methods across a hyperparameter sweep (via random search for each method) under heterogeneous settings. Results are shown for (a) ResNet-18 on CIFAR-10, (b) ResNet-50 on CIFAR-100.}
\label{fig:test}
\end{figure}

\textbf{Usable Rate of Federated Learning Methods:} A key advantage of a hyperparameter-free method is its ability to be deployed without additional engineering effort. To demonstrate practical deployability, we examine the number of usable models, defined as those achieving final accuracy above a deployment threshold of 80\% of the highest classification accuracy. Table \ref{tab:useable_models} reports the percentage of hyperparameter configurations that produced usable models for each method.

In all experiments, Adaptive FedECADO achieves a usable rate of 100\%, reliably producing high-performance models regardless of hyperparameter choices. In contrast, comparison methods often result in low-performance models that consume significant computational resources during training.

\textbf{Perturbing Hyperparameter Selections:}
We also evaluate each method’s sensitivity to hyperparameter perturbations by varying optimal values by 20\% and measuring the impact on final accuracy (Appendix~\ref{app:perturbing_hyperparameters}). Adaptive FedECADO remains stable and performant across all perturbations, showing robustness to its single hyperparameter, $\gamma$. In contrast, methods like FedAdam and FedAdaGrad exhibit significant performance drops, underscoring their dependence on precise tuning.

\paragraph{Perturbing Hyperparameter Choices in Adaptive FedECADO}
We further evaluate the convergence behavior of Adaptive FedECADO under varying values of $\gamma$. Figure \ref{fig:eta_sweep} shows training trajectories on the CIFAR-10 dataset using a ResNet-18 model across four orders of magnitude of $\gamma$, demonstrating robustness to its sole hyperparameter.

\paragraph{Adding Schedulers to Client Updates} Schedulers can be added to client updates; however, they introduce additional hyperparameters can improve stability at the cost of damping convergence speed. The impact of client-side schedulers to FedAdam and FedAdaGrad is shown in Appendix \ref{app:schedulers_experiment}.
\section{Conclusion}

We introduce a fully adaptive federated learning method that eliminates the need for hyperparameter tuning in heterogeneous settings. By modeling federated learning as a dynamical system, our approach derives adaptive momentum and learning rates using principles from critical damping and adaptive simulation. The result is a momentum-based algorithm that achieves fast, stable convergence across non-IID data distributions and varying client compute, all controlled by a single global hyperparameter. We show that our method is highly robust to the choice of this hyperparameter. Compared to existing adaptive methods, Adaptive FedECADO delivers higher classification accuracy without tuning, whereas other methods are highly sensitive to hyperparameter selection. This makes our approach particularly well-suited for rapid prototyping by lowering the engineering overhead typically required to implement federated learning in heterogeneous environments.

\newpage 

\bibliographystyle{plain}
\bibliography{main}

\clearpage
\appendix

\section{Equivalent Circuit Model of Federated Learning}
\label{app:circuit_background}

Federated learning is a decentralized optimization framework in which a central server coordinates the training of a global model using gradient updates computed independently on distributed client devices. The global model parameters, $x_k$ at iteration \( k \) are updated by gradient descent:
\begin{equation}
    x_{k+1} = x_k - \alpha \sum_{i=1}^{N} \nabla f_i(x_k),
\end{equation}
where \( \nabla f_i(x_k) \) is the gradient of the local objective function for client \( i \), and \( \alpha \) is a global learning rate. As \( \alpha \to 0 \), this update can be represented as a time-discretized version of a continuous-time dynamical system:
\begin{equation}
    \dot{x}(t) = -\sum_{i=1}^{N} \nabla f_i(x(t)),
    \label{eq:gradient_flow}
\end{equation}
where \( \dot{x}(t) \) denotes the time derivative of the model parameters. While the continuous-time model in \eqref{eq:gradient_flow} offers useful insight into the dynamics of the optimization variables, it has a key limitation in the federated setting: all local clients use the global states, \( x(t) \), to compute the local updates $\nabla f_i(x)$, however, in the federated setting, clients compute local updates using their own local states.

To better represent the structure of federated learning, \cite{agarwal2023equivalent} introduced a continuous-time model that separates the global state from the individual client states using auxiliary flow vector, $\ILi$. In this formulation, the global model, $x_c(t)$, is maintained by the central agent, and each client \( i \) maintains its own local state denoted as \( x_i(t) \). The interaction between a client and the server is captured through the auxiliary vector \( \ILi(t) \). This results in the following dynamical system:
\begin{align}
\label{eq:appA_1}
    \frac{d}{dt}x_c(t) + \sum_{i=1}^{|\mathcal{C}|} \ILi(t) = 0, \\\
    \label{eq:appA_2}
    L_i \dot{\ILi}(t) = x_c(t) - x_i(t) \quad \forall i \in [1,|\mathcal{C}|], \\
    -\ILi(t) + \frac{d}{dt}x_i(t) + p_i \nabla f_i(x_i(t)) = 0 \quad \forall i \in [1,|\mathcal{C}|],
    \label{eq:appA_3}
\end{align}
where \( L_i \) is a scalar parameter that determines the coupling between client \( i \) and the central agent.

\begin{figure}
    \centering
    \includegraphics[width=0.7\linewidth]{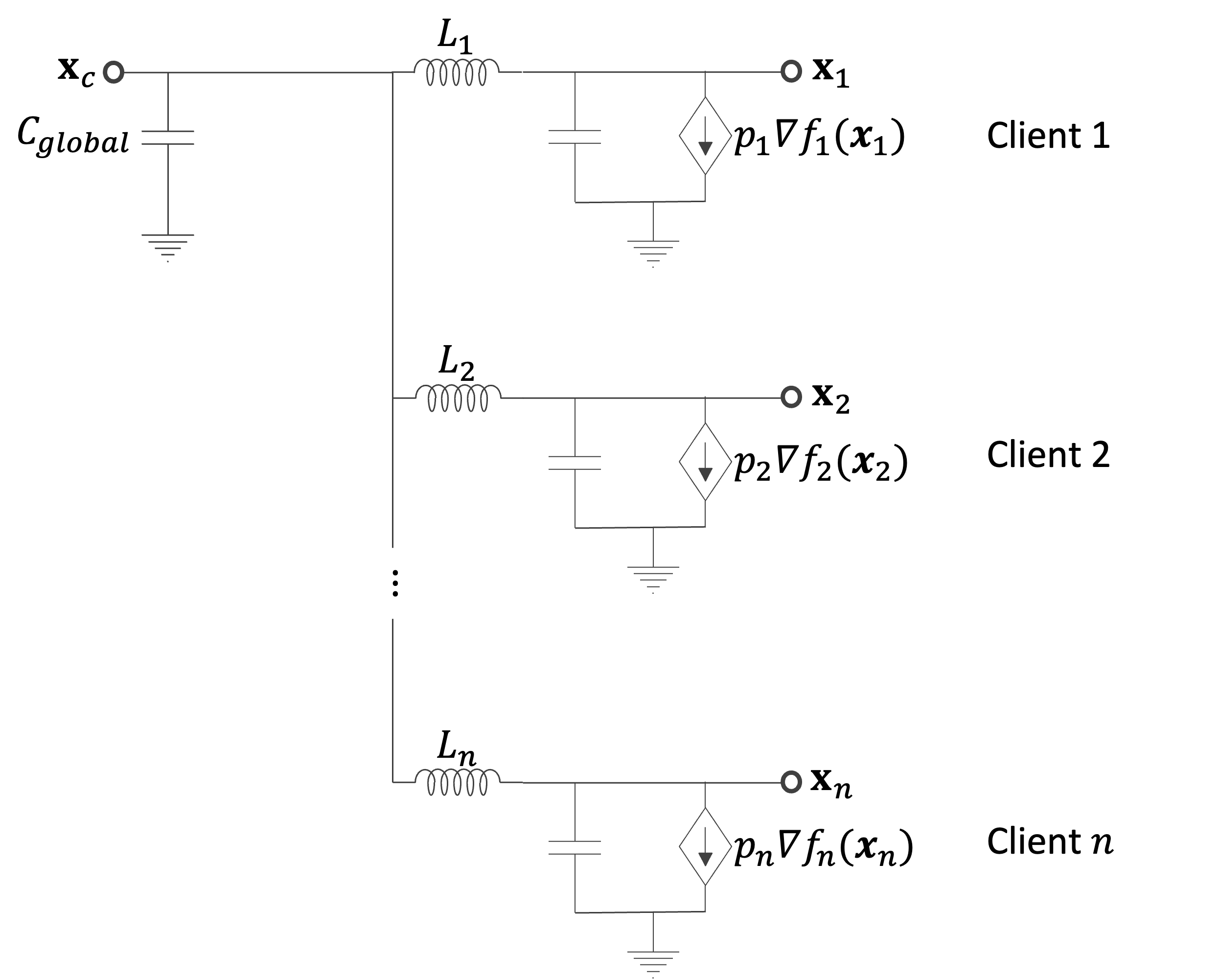}
    \caption{The dynamical system model of federated learning can be represented by an electrical circuit. In the circuit representation, each global state, $x_c$, is modeled by a node-voltage connected to multiple nodes whose voltage represents the client states, $x_i$. These nodes are connected by an inductor whose dynamics are modeled by \eqref{eq:inductor_ode}.}
    \label{fig:equivalent_circuit}
\end{figure}

In this work, we model the dynamical system in \eqref{eq:appA_1}-\eqref{eq:appA_3} as electrical circuits to leverage physical analysis and simulation techniques to design hyperparameters. Specifically, the dynamical system can be represented by the circuit in Figure \ref{fig:equivalent_circuit} which mirrors the structure of the federated learning process. In this circuit, the central server states are modeled as voltages at a central node, \( x_c(t) \), connected to node whose voltages represent the client state-variables, \( x_i(t) \). These nodes are connected by flow variable \( \ILi(t) \) that represents the electrical current between the client and the central node. 

In this circuit-based model, each client node is connected to a capacitor, which models the continuous-time dynamics of each client’s state. A capacitor is an electrical circuit component that stores energy in an electric field and whose voltage changes based on the net current flowing into it. Mathematically, the current-voltage behavior of a capacitor is described by:
\begin{equation}
    I_C = C \dot{V},
\end{equation}
where \( I_C \) is the current, \( C \) is the capacitance, and \( \dot{V} \) is the rate of voltage change. In our analogy, the voltage across this capacitor corresponds to the local client states, $x_i(t)$. Each client node is connected to a voltage-controlled current source, which outputs a current equal to the local gradient \(p_i \nabla f_i(x_i(t)) \). 

The client nodes are connected to the central node via inductors. An inductor is another basic circuit element that stores energy in a magnetic field and resists changes in current. It follows the current-voltage relation:
\begin{equation}
    V_L = L \dot{I},
\end{equation}
where \( V_L \) is the voltage across the inductor, \( L \) is the inductance, and \( I \) is the current. In our model, the voltage across the inductor is the difference between the global and client states, \( x_c(t) - x_i(t) \), and the resulting current is \( \ILi(t) \). This implies that the inductor accumulates the difference between the global and local models over time, scaled by the inductance parameter \( L_i \). Adding the inductors  introduces a momentum-like effect into the client-server interaction.

 Together, these components form a complete physical circuit, whose behavior is governed by Kirchhoff’s Current Law (KCL). KCL is a fundamental principle in circuit theory which states that the total current entering a node must equal the total current leaving it.

Applying KCL at the central node yields the equation \( \dot{x}_c(t) = -\sum_i \ILi(t) \), representing the accumulation of all client currents into the global capacitor. 

Applying KCL at each client node, the inductor current flowing into each client equals the sum of the capacitor current and the gradient-induced current. This equates to the client ODE $ -\ILi(t) + \frac{d}{dt}x_i(t) + p_i \nabla f_i(x_i(t)) = 0$.

By modeling the federated learning system using these physical components, we obtain a circuit-based representation that is structurally identical to the federated learning process. This perspective enables us to derive adaptive step sizes and momentum terms directly from physical principles, rather than relying on empirical tuning or heuristic modifications. 

\subsection{Modeling the Central Agent Aggregation as a Circuit}

\begin{figure}
    \centering
    \includegraphics[width=0.5\linewidth]{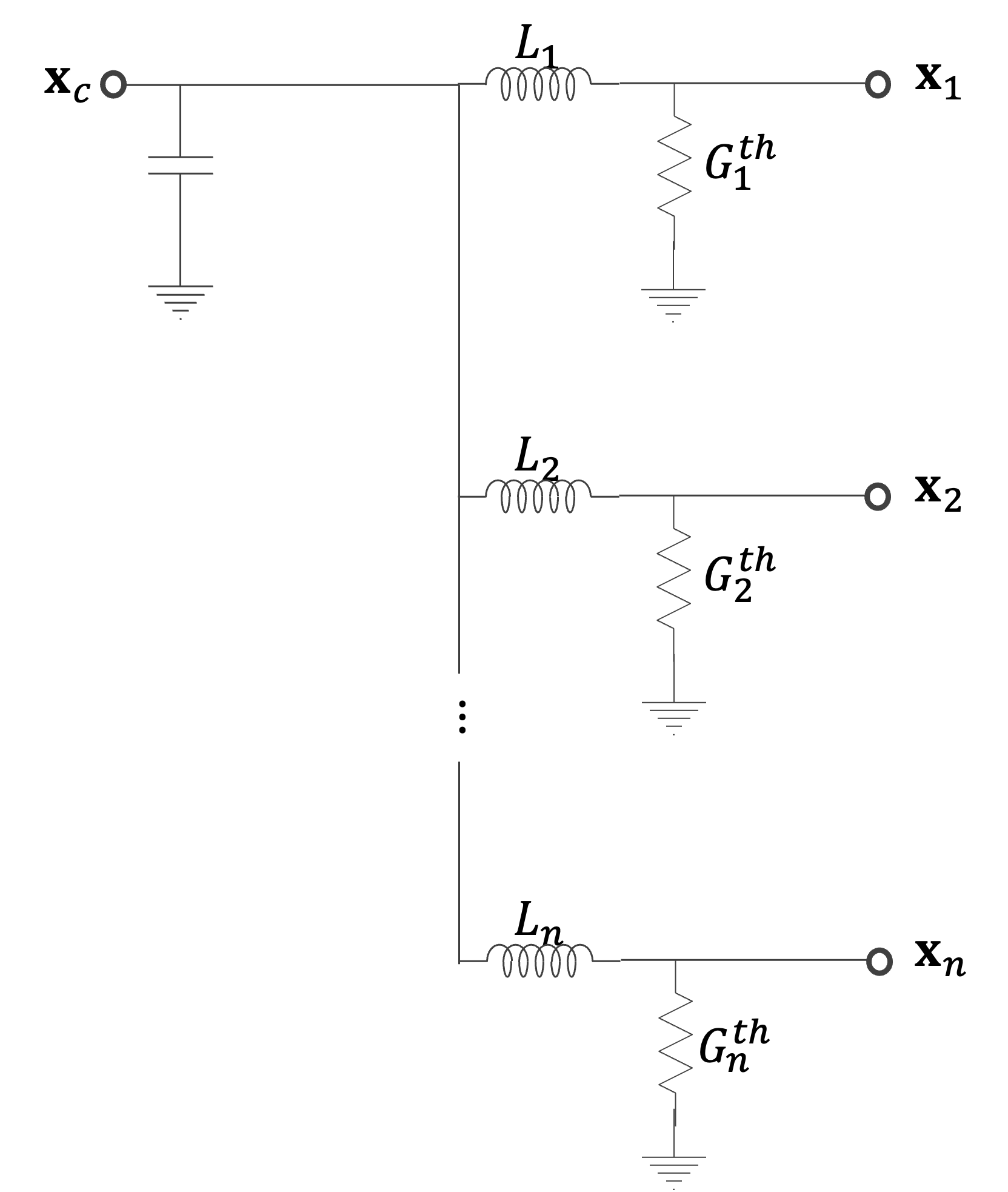}
    \caption{Equivalent circuit represented of the aggregation step for central agents in \eqref{eq:central_agent_capacitor_gs}-\eqref{eq:central_agent_inductor_gs}}
    \label{fig:aggregation_linear_circuit}
\end{figure}

 The continuous-time dynamics at aggregation round \( k+1 \), which can be described by the following system of equations:

\begin{align}
    \dot{x}_c^{k+1}(t) &= \sum_{i=1}^n \ILi^{{k+1}}(t),
    \label{eq:central_agent_capacitor_gs} \\
    L \dot{\ILi}^{{k+1}}(t) &= x_c^{k+1}(t) - \left(\ILi^{{k+1}}(t)\hat{G}_i^{\text{th}^{-1}} + x_i^{k+1}(t) - \ILi^{{k}}(t)\hat{G}_i^{\text{th}^{-1}}\right),
    \label{eq:central_agent_inductor_gs}
\end{align}

where \( \hat{G}_i^{\text{th}} \) represents a linear sensitivity of each client, and \( L_i \) is the momentum term for each client branch. This set of linear ODEs can be modeled by an electrical circuit shown in Figure \ref{fig:aggregation_linear_circuit}.

This circuit differs from the previous nonlinear circuit representation of Figure \ref{fig:equivalent_circuit} in that the client models are linearized via a first-order approximation using the sensitivity matrix \( \hat{G}_i^{\text{th}} \). Physically, the sensitivity term, \( \hat{G}_i^{\text{th}} \), behaves like a \textbf{resistor} in the circuit, which relates voltage and current linearly through Ohm’s Law. This linear sensitivity is derived using a circuit concept known as \textbf{Thevenin impedance}, which models the sensitivity of current of a circuit network with respect to the node-voltage as an impedance and is used to simplify the analysis of linear electrical networks.


\subsubsection{Client Perspective via Thevenin Impedance}

The linear circuit in Figure \ref{fig:aggregation_linear_circuit} presents a complex interconnected network where the clients and central server states are coupled. The coupled interactions makes it challenging to derive momentum term, $L_i$, that makes the entire linear circuit critically damped. 

Instead, we study the behavior of each individual client during aggregation using a \textit{Thevenin impedance} looking outward from a single client branch into the rest of the circuit. From the client’s perspective, this impedance includes the contribution of the central agent capacitor as well as all the other clients’ branches connected in parallel. In our experiments we find that this Thevenin impedance is dominated by the central agent’s capacitor, due to its relatively large capacitance compared to the resistance and inductance in the client branches.

This observation allows us to approximate the Thevenin impedance looking out of the client branch as the central agent capacitor. As a result, each client branch looks like a \textbf{series RLC circuit}, consisting of:
\begin{itemize}
    \item a resistor \( R = \hat{G}_i^{\text{th}^{-1}} \), representing the client’s linearized sensitivity,
    \item an inductor \( L_i \), representing the momentum term, and
    \item a capacitor \( C_{global} =1\), representing the central agent’s state.
\end{itemize}

\begin{figure}
    \centering
    \includegraphics[width=0.5\linewidth]{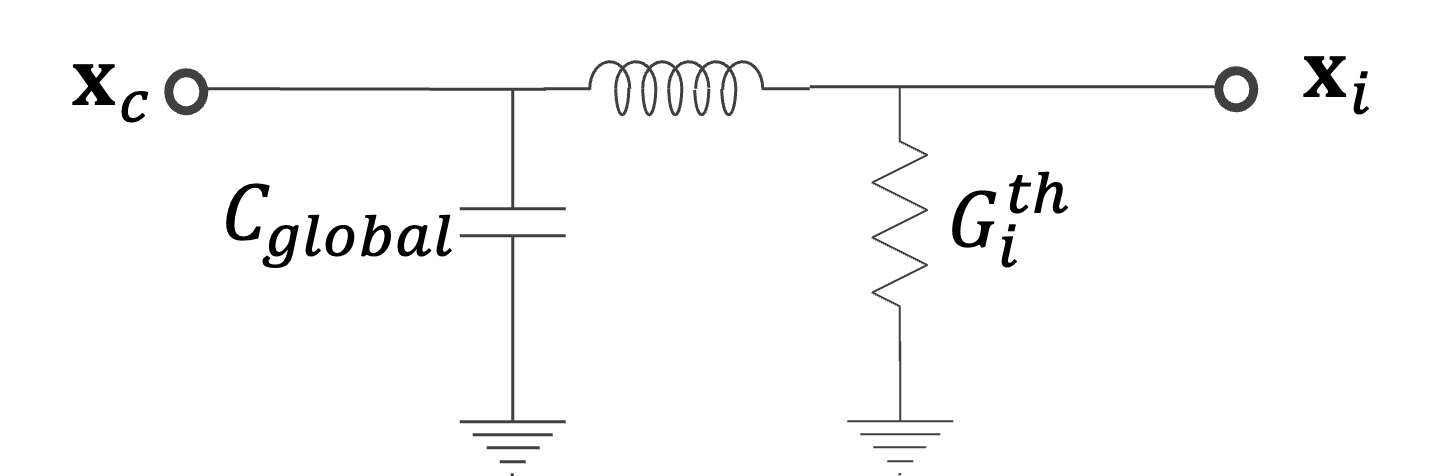}
    \caption{Each client branch in the aggregation circuit (Figure \ref{fig:aggregation_linear_circuit}) can be reduced to a series RLC since the Thevenin impedance looking out of each client branch effectively looks like a capacitance for the central agent.}
    \label{fig:series_rlc}
\end{figure}

This reduced model is depicted in Figure~\ref{fig:series_rlc}. The series RLC circuit abstraction provides a highly interpretable local model for each client, enabling us to analyze convergence, damping, and oscillation behavior using well-known results from second-order systems. By modeling each client-server interaction as a damped oscillator, we can derive stability conditions and design update rules with predictable behavior, without relying on heuristic hyperparameter tuning.

\subsection{Series RLC Behavior and Designing for Critical Damping}
\label{app:critically_damped_L}
Each client-server connection in our federated learning framework can be modeled as a series RLC circuit.
The behavior of this circuit is governed by a second-order differential equation that can be damped by tuning the inductance, $L_i$, as shown by Figure \ref{fig:simple_rlc_damping}. Ideally, we want the system to be critically damped, meaning it returns to equilibrium as quickly as possible without oscillation. In the context of federated optimization, this corresponds to designing a momentum term for the aggregation step that achieves fast global convergence without overshooting.

To achieve critical damping in a series RLC circuit, we must design the inductance \( L_i \) relative to the client sensitivity, $\hat{G}_i^{th}$, and the capacitance \( C_{global}=1 \). We begin by recalling that the differential equation governing the series RLC circuit is:

\begin{equation}
    L_i \ddot{q}(t) + R \dot{q}(t) + \frac{1}{C} q(t) = 0,
\end{equation}

where \( q(t) \) is the charge on the capacitor, \( \dot{q}(t) \) is the capacitor current , \( L_i \) is the inductance, \( R=\hat{G}_i^{{th}^{-1}} \) is the resistance, and \( C=1 \) is the capacitance. This equation has the same form as a damped second-order linear system and is derived in circuit literature \cite{agarwal2005foundations}. From this, we can derive a damping ratio \( \zeta \), which determines whether the system is underdamped, critically damped, or overdamped, is given by:

\begin{equation}
    \zeta = \frac{R}{2} \sqrt{\frac{C}{L_i}}.
\end{equation}

Critical damping occurs when \( \zeta = 1 \), leading to the condition:

\begin{equation}
    1 = \frac{R}{2} \sqrt{\frac{C}{L_i}}.
\end{equation}

Solving for \( L_i \), we square both sides:

\begin{equation}
    1 = \frac{R^2 C}{4L_i} \quad \Rightarrow \quad L_i = \frac{C R^2}{4}.
\end{equation}

Thus, to achieve critical damping, the inductance must be set according to:

\begin{equation}
    L_i = \frac{C R^2}{4} = \frac{1}{4} \hat{G}_i^{{th}^{-2}}.
\end{equation}

This equation gives us a principled way to design the inductance (i.e., momentum term) based on the client sensitivity \(  \hat{G}_i^{{th}} \). Rather than relying on heuristics or manual tuning, we can compute the inductance directly to ensure optimal convergence behavior. In practice, this means that each client can be configured with a momentum that improves convergence speed in the global learning process.
\section{Derivation of Local Truncation Error for Forward-Euler}
\label{app:lte_fe_derivation}

To numerically solve the client ODE in \eqref{eq:client_ode_gs}, we apply the Forward Euler method with time step \( \Delta t \). The discretized update rule is:

\begin{equation}
x_i(t+\Delta t) = x_i(t) - \Delta t \left( \nabla f(x_i(t)) + \ILi(t) \right),
\end{equation}

The local truncation error (LTE) measures the error introduced in a single step of the numerical method, defined as:

\begin{equation}
\varepsilon_{FE} = \frac{1}{\Delta t} \left[ x_i(t+\Delta t) - x_i(t) + \Delta t \left( \nabla f(x_i(t)) + \ILi(t) \right) \right].
\end{equation}

To compute this, we expand \( x_i(t+\Delta t) \) using a Taylor series about \( t \):

\begin{equation}
x_i(t+\Delta t) = x_i(t) + \Delta t \dot{x}_i(t) + \frac{\Delta t^2}{2} \ddot{x}_i(t) + \mathcal{O}(\Delta t^3).
\end{equation}

From the ODE \eqref{eq:client_ode_gs}, we know:

\begin{equation}
\dot{x}_i(t) = -\nabla f(x_i(t)) - \ILi(t).
\end{equation}

Substituting into the expansion:

\begin{align}
x_i(t+\Delta t) &= x_i(t) - \Delta t \left( \nabla f(x_i(t)) + \ILi(t) \right) + \frac{\Delta t^2}{2} \ddot{x}_i(t) + \mathcal{O}(\Delta t^3).
\end{align}

Now, substituting into the expression for \( \varepsilon_{FE} \):

\begin{align}
\varepsilon_{FE_n} &= \frac{1}{\Delta t} \left[ - \Delta t \left( \nabla f(x_i(t)) + \ILi(t) \right) + \frac{\Delta t^2}{2} \ddot{x}_i(t) + \mathcal{O}(\Delta t^3) + \Delta t \left( \nabla f(x_i(t)) +\ILi(t) \right) \right] \nonumber \\
&= \frac{1}{\Delta t} \left[ \frac{\Delta t^2}{2} \ddot{x}_i(t) + \mathcal{O}(\Delta t^3) \right] \nonumber \\
&= \frac{\Delta t}{2} \ddot{x}_i(t) + \mathcal{O}(\Delta t^2).
\end{align}

In the expression, $\ddot{x}_i(t)$ can be evaluated as:
\begin{equation}
   \ddot{x}_i(t) = \ILi(t) + \nabla f_i(x_i(t)) - \ILi(t-\Delta t) - \nabla f_i(x_i(t-\Delta t)) .
\end{equation}
Using this expression, the LTE for Forward-Euler is approximated by the first-order term as:

\begin{equation}
    \varepsilon_{FE_n} = \frac{\Delta t}{2} \left[ \ILi(t) + \nabla f_i(x_i(t)) - \ILi(t-\Delta t) - \nabla f_i(x_i(t-\Delta t)) \right].
\end{equation}

\section{Derivation for Backward-Euler Accuracy and Stability}
\label{app:be_derivation}

The aggregation of client updates is modeled by a linear set of ODEs in \eqref{eq:central_agent_capacitor_gs}-\eqref{eq:central_agent_inductor_gs}. To numerically determine the central agent state, we discretize these ODEs using the Backward Euler method with a time step of \( \Delta t \).  The numerical update is:

\begin{equation}
x_c^{k+1}(\tau+\Delta t) = x_c^{k+1}(\tau) - \Delta t \sum_{i=1}^{n} \ILi^{{k+1}}(\tau+\Delta t),
\end{equation}
\begin{equation}
\begin{aligned}
\ILi^{{k+1}}(\tau+\Delta t) = \ILi^{{k+1}}(\tau) + \frac{\Delta t}{L} \Big( x_c^{k+1}(\tau+\Delta t) - ( \ILi^{{k+1}}(\tau+\Delta t) G_i^{th^{-1}} \
\\+ \Gamma(x_{i}^{k+1}(t),\tau+\Delta t) - \ILi^{k}(\tau+\Delta t) G_i^{th^{-1}} ) \Big).
\end{aligned}
\end{equation}

To compute the local truncation error, we assume the previous value is exact and define the error for the first equation as:

\begin{equation}
\varepsilon_{BE}^C = \frac{1}{\Delta t} \left( x_c^{k+1}(t+\Delta t) - x_c^{k+1}(t) - \Delta t \sum_i \ILi^{k+1}(t+\Delta t) \right),
\end{equation}

and for the second equation:

\begin{align}
\varepsilon_{BE}^L &= \frac{1}{\Delta t} \left( L_i (\ILi^{k+1}(t+\Delta t) - \ILi^{k+1}(t)) \right. \nonumber \\
&\quad \left. - \Delta t \left[ x_c^{k+1}(t+\Delta t) - \left( \ILi^{k+1}(t+\Delta t) G_i^{\text{th}^{-1}} + \Gamma(x_i^{k+1}(t+\Delta t), \Delta t) - \ILi^{k} G_i^{\text{th}^{-1}} \right) \right] \right).
\end{align}

Expanding the exact solutions using Taylor series:

\begin{align}
x_c^{k+1}(t+\Delta t) &= x_c^{k+1}(t) + \Delta t \dot{x}_c^{k+1}(t) + \frac{(\Delta t)^2}{2} \ddot{x}_c^{k+1}(t) + \mathcal{O}((\Delta t)^3), \\
\ILi^{k+1}(t+\Delta t) &= \ILi^{k+1}(t) + \Delta t \dot{\ILi}^{k+1}(t) + \frac{(\Delta t)^2}{2} \ddot{\ILi}^{k+1}(t) + \mathcal{O}((\Delta t)^3).
\end{align}

Substituting into \( \varepsilon_{BE}^C \):

\begin{align}
\varepsilon_{BE}^C &= \frac{1}{\Delta t} \left( \Delta t \dot{x}_c^{k+1}(t) + \frac{(\Delta t)^2}{2} \ddot{x}_c^{k+1}(t) - \Delta t \sum_i \ILi^{k+1}(t) - (\Delta t)^2 \sum_i \dot{\ILi}^{k+1}(t) + \mathcal{O}((\Delta t)^3) \right) \nonumber \\
&= \left( \dot{x}_c^{k+1}(t) - \sum_i \ILi^{k+1}(t) \right) + \mathcal{O}(\Delta t).
\end{align}

Using the ODE definition, \( \dot{x}_c^{k+1}(t) = \sum_i \ILi^{k+1}(t) \), so:

\begin{equation}
\tau_n^{(x)} = \mathcal{O}(\Delta t).
\end{equation}

A similar analysis for \( \tau_n^{(\ILi)} \) shows that it is also \( \mathcal{O}(\Delta t) \). Therefore, Backward Euler remains first-order accurate when applied to this linear system.

\subsection{Numerical Stability of Backward Euler}
The Backward-Euler integration is a numerically stable algorithm, which means for a linear system, we can guarantee convergence to the steady-state (or stationary point) regardless of the choice of $\Delta t$.
To analyze property of numerical stability, we examine the linear system:

\begin{equation}
\dot{y}(t) = A y(t),
\end{equation}
where $A \succ0$ is a linear system.
The Backward Euler update is:

\begin{equation}
y_{n+1} = (I-\Delta t A)^{-1}y_n.
\end{equation}

This scheme is stable if \( eig(I-\Delta t A)^{-1} \leq 1 \). For all \( A\succ0 \), the value of $eig(I-\Delta t A)^{-1} $ is less than one. This implies that Backward Euler is \textbf{A-stable}: it remains stable for all step sizes \( \Delta t > 0 \),   in the presence of stiff dynamics.

In the context of our aggregation ODE system, where the linear dynamics are governed by parameters \( G_i^{\text{th}} \succ 0 \) and \( L_i > 0 \), the eigenvalues of the linearized operator have negative real parts. Therefore, the Backward Euler method guarantees stability regardless of the time step \( \Delta t \), making it a robust choice for simulating the dynamics of client-server interactions in heterogeneous and stiff federated learning systems.
\section{Derivation of  Damping Condition}
\label{app:max_eig_derivation}

Consider the linear set of ODEs representing the central agent aggregation \eqref{eq:central_agent_capacitor_gs}-\eqref{eq:central_agent_inductor_gs}. We abstract the linear ODEs to the form:

\begin{equation}
C \dot{x}(t) = G x(t),
\label{eq:linear_cg_system}
\end{equation}

where \( x(t)=\begin{bmatrix}
    x_c(t) \\ \IL(t)
\end{bmatrix} \) , \( C = \begin{bmatrix}
    \mathcal{I}, 0 \\ 0, L
\end{bmatrix} \) is a symmetric positive definite matrix representing capacitance and inductances.  \( G =\begin{bmatrix}
    0, 0,  ... \\ 0, \hat{G}_i^{th}, 0 \\
    0,0, \ddots
\end{bmatrix} \) is a positive semi-definite matrix representing client sensitivities.

Rewriting this system in standard form gives:

\begin{equation}
\dot{x}(t) = C^{-1} G x(t).
\end{equation}

The dynamics of this system are governed by the eigenvalues of the matrix \( C^{-1} G \). Let \( \lambda_i \in \mathbb{C} \) denote the eigenvalues of \( C^{-1} G \). These eigenvalues determine the transient behavior of the system: real negative eigenvalues lead to exponential decay, while complex eigenvalues with nonzero imaginary parts introduce oscillations.

In control theory, a system is said to be \textbf{critically damped} when it returns to equilibrium as quickly as possible without oscillation. For a linear system, this corresponds to all eigenvalues of the system matrix being real and negative.

In the context of Equation~\eqref{eq:linear_cg_system}, achieving critical damping involves two primary goals:

\begin{enumerate}
    \item \textbf{Ensure all eigenvalues of \( C^{-1} G \) are real and non-positive.} This removes imaginary components to avoid oscillations in the solution trajectories.
    \item \textbf{Maximize the smallest (least negative) real eigenvalue of \( C^{-1} G \).} This improves the convergence rate of the slowest mode and ensures the system returns to equilibrium quickly and uniformly.
\end{enumerate}

Formally, if \( \Lambda(C^{-1} G) = \{\lambda_1, \ldots, \lambda_d\} \subset \mathbb{R} \) denotes the eigenvalues of \( C^{-1} G \), we aim to solve:

\begin{equation}
\max_{L \succ 0} \quad \inf_{i} \ \text{Re}(\lambda_i(C^{-1} G)) \quad \text{subject to} \quad \text{Im}(\lambda_i(C^{-1} G)) = 0, \ \forall i.
\label{eq:critical_damping_optimization}
\end{equation}

This objective seeks the choice of inductance \( L \) that spreads the eigenvalues of \( C^{-1} G \) onto the negative real axis and pushes smallest \( |\lambda_i| \) as far left as possible.

\section{Adaptive Step Size Selection for Client Steps}
\label{app:fe_adaptive_time_step}

The adaptive step-size selection procedure for an individual client is outlined in Algorithm \ref{fe_adaptive_time_step}. The algorithm is initialized with a time step $\Delta t$ based on the passivity conditions defined in Equation \eqref{tfe-naive}. A backtracking line search (line 4) is then employed to iteratively adjust the step size, ensuring that the local truncation error of the Forward-Euler integration for each client’s ODE remains within the prescribed global tolerance $\gamma$ (line 3). The backtracking line-search damps the time-step according to the ratio, $0<\frac{\gamma}{\max|\varepsilon_{FE}|}<1$, which scales the step-size by how far the local truncation error is from the tolerance, $\gamma$.

\begin{algorithm}
\caption{Adaptive Step Size Selection for Client Step}
\label{fe_adaptive_time_step}
\begin{algorithmic}[1]
\STATE{\textbf{Input: }$x_i^{k+1}(t)$, $\ILi^k(t)$,$x_c^{k+1}(t)$, $\gamma>0$}
\STATE{$\Delta t \gets Equation $\eqref{tfe-naive}}
\WHILE{$\max|\varepsilon_{FE}| \geq \gamma$}
\STATE{$\Delta t = \frac{\gamma}{\max|\varepsilon_{FE}|} \Delta t$}
\STATE{$\varepsilon_{FE} \gets$ Equation \eqref{eq:client_ode_fe}($\Delta t$)}
\ENDWHILE
\RETURN $\Delta t$ 
\end{algorithmic}
\end{algorithm}
\section{Adaptive Step Size Selection for Central Agent Aggregation}
\label{app:be_adaptive_time_step}

The adaptive step-size selection of the central agent aggregation is shown in Algorithm \ref{be_adaptive_time_step}. This algorithm uses a backtracking line-search in line 4 to adaptive select a step-size that ensures the local truncation error of the Backward-Euler integration of the central agent ODEs are within a global tolerance, $\gamma$ (line 3).

\begin{algorithm}
\caption{Adaptive Step Size Selection for Central Agent Aggregation}
\label{be_adaptive_time_step}
\begin{algorithmic}[1]
\STATE{\textbf{Input: }$x_c^{k+1}(t)$, $\ILi^k(t)$,$x_i^{k+1}(t)$, $\gamma>0$, $\Delta t_k$}
\STATE{$\Delta t \gets (\Delta t)_k$}
\WHILE{$\max|\varepsilon_{BE}| \geq \gamma$}
\STATE{$\Delta t = \frac{\gamma}{\max|\varepsilon_{BE}|} \Delta t$}
\STATE{$\varepsilon_{BE} \gets$ \eqref{eq:be_lte_cap}, \eqref{eq:be_lte_ind}}
\ENDWHILE
\RETURN $\Delta t$ 
\end{algorithmic}
\end{algorithm}
\section{Adaptive Federated Learning Algorithm}
\label{app:adaptive_fedecado_algo}

The workflow for Adaptive FedECADO is shown in Algorithm \ref{adaptive_fedecado_alg}. The input to the algorithm include each client's loss function and a global, scalar hyperparamter, $\gamma>0$ that controls the local truncation errors for the central agent aggregation and all client simulations. 

Prior to beginning the client communication, Adaptive FedECADO computes each client's sensitivity in line 5 as well as the corresponding momentum parameters in line 6. Then, a subset of active clients perform local updates using a Forward-Euler integration as shown in line 13, where the local step sizes adapt according to Algorithm \ref{fe_adaptive_time_step}. Each active client then communicates the final state vector and simulation window in line 14 to the central server. The aggregation of the client updates are then performed in line 18-19 according to a Backward-Euler integration of the central agent ODEs. The step-sizes for the central agent are determined using Algorithm \ref{be_adaptive_time_step}, which is designed to control the local truncation error within a predefined tolerance, $\gamma$.

\begin{algorithm}[H]
    \caption{Adaptive FedECADO Algorithm}
    \label{adaptive_fedecado_alg}
    \textbf{Input: } $\nabla f_i(\cdot)$,$x(0)$, $\gamma>0$
    
    \begin{algorithmic}[1]
    \STATE{$x_c \gets x(0)$}
    \STATE{$x_i \gets x(0)$}
    \STATE{$I_i^L \gets 0$}
    \STATE{$t \gets 0$}

    \STATE{Precompute $\bar{G}_i^{th} \; \forall i\in C$ via \eqref{eq:constant_sensitivity_def}} 
    
    \STATE{Precompute $L_i \; \forall i\in C $ via \eqref{eq:critical_L}}
    
    \STATE{\textbf{do while} $\|\dot{x}_c\|^2 > 0$}
    \STATE{\hspace*{\algorithmicindent}$x_c^k \gets x_c^{k+1}$}
    \STATE{\hspace*{\algorithmicindent}$x_i^{k} \gets x^{i^{k+1}}$}
    
    \STATE{\hspace*{\algorithmicindent}\textit{Parallel Solve for active client states, $x_i^{k+1}(t+T_i) \forall i\in C_a$, by simulating:} }

   \STATE{\hspace*{\algorithmicindent} \hspace*{\algorithmicindent}\textbf{for $e_i$ epochs:}}

      \STATE{\hspace*{\algorithmicindent}\hspace*{\algorithmicindent}\hspace*{\algorithmicindent}\hspace*{\algorithmicindent}$\Delta t_i \gets$ Algorithm \ref{fe_adaptive_time_step}} 
    \STATE{\hspace*{\algorithmicindent}\hspace*{\algorithmicindent}\hspace*{\algorithmicindent}\hspace*{\algorithmicindent}$x_i^{k+1}(t+\Delta t_i) = x_i^{k+1}(t) - \Delta t_i\nabla f(x_i^{k+1}(t)) - \Delta t_i I_i^{L^k}(t) $}

      \STATE{\hspace*{\algorithmicindent}\hspace*{\algorithmicindent} Communicate $x_i(T_i)$ and $T_i$ to central server} 
    
    \STATE{\hspace*{\algorithmicindent}\textit{Solve central agent aggregation by simulating:} }

    \STATE{\hspace*{\algorithmicindent} \textbf{for $\tau \in[t,t+\max(T_i)]$}}
    
    \STATE{\hspace*{\algorithmicindent} \hspace*{\algorithmicindent}Select $\Delta t$ according to Algorithm \ref{be_adaptive_time_step}}
     \STATE{\hspace*{\algorithmicindent}\hspace*{\algorithmicindent}  Evaluate active client states at timepoint $\tau$: $\Gamma(x_i^{k+1},\tau) \;\;\forall i \in C_a$}
    
         \STATE{\hspace*{\algorithmicindent}\hspace*{\algorithmicindent}\textit{Solve for $x_c^{k+1}(\tau+\Delta t), I_i^{L^{k+1}}(\tau+\Delta t)$} according to \eqref{eq:central_agent_be_cap}-\eqref{eq:central_agent_be_ind}
}

    \STATE{\hspace*{\algorithmicindent}\hspace*{\algorithmicindent} $\tau = \tau+\Delta t$}
    \STATE{Return $x_c$}
    \end{algorithmic}
    \end{algorithm}
\section{Estimating Critical Damping}
\label{app:critically_damped_L}

In Section \ref{sec:momentum_params}, we develop an approximate solution to \eqref{eq:critical_damping_eig} that relies on the structure of the analog circuit model of federated learning. The approximation is developed by concluding that the sensitivity of $\ILi$ with respect to the global state variable, $x_c$, is mainly attributed to the dynamics of the global state, $\frac{d}{d x_c} \dot{x}_x$. 

We can demonstrate this numerically by training ResNet-18 model on CIFAR-10 dataset through the federated setting across 3 clients. In this experiment, we create the linearized aggregation model in \eqref{eq:central_agent_capacitor_gs}-\eqref{eq:central_agent_inductor_gs}. Then we study the sensitivity of the flow variable of the first client, $I_{L_1}$, due to perturbations in the central agent states, $x_c$, and the flow variables in the other clients, $I_{L_2}$ and $I_{L_3}$. Table \ref{tab:sensitivity_three_client} demonstrates the normalized sensitivities across all state-variables.
We observe that the dominant sensitivity is due to $x_c$ as opposed to changes in the other client's flow variables. This experimentally justifies that the dominant sensitivity term of any client is due to changes in the central agent state, thus allowing us to make an appropriate approximation in Section \ref{sec:momentum_params}. 

\begin{table}[h!]
    \centering
    \begin{tabular}{|c|c|}
    \hline
       Perturbation Variable  & Normalized Sensitivity $\partial I_{L_1}$ \\\hline
        $x_c$ & 0.97 \\\hline
        $I_{L_2}$ & 0.014 \\\hline
        $I_{L_3}$ & 0.016 \\\hline        
    \end{tabular}
    \caption{Sensitivity analysis of a single client flow vector $I_{L_1}$ in the central agent aggregation dynamics \eqref{eq:central_agent_capacitor_gs}–\eqref{eq:central_agent_inductor_gs}. We perturb the central agent state $x_c$ and the flow variables of clients 2 and 3, $I_{L_2}$ and $I_{L_3}$, and measure the resulting change in $I_{L_1}$. Sensitivities are computed as partial derivatives and normalized by the sum of all absolute sensitivities: $| \frac{\partial I_{L_1}}{\partial x_c}| + | \frac{\partial I_{L_1}}{\partial I_{L_2}}| + | \frac{\partial I_{L_1}}{\partial I_{L_3}}|$. The table reports the normalized sensitivity of $I_{L_1}$ to each perturbed variable.}
    \label{tab:sensitivity_three_client}
\end{table}
\section{Varying Hyperparameters in Federated Learning Methods}
\label{app:hyperparameter_ranges}

We vary each federated learning method's hyperparameters to study their sensitivity to the model performance. The hyperparameters are randomly selected within the bounds listed in the following tables.

\begin{table}[h!]
    \centering
    \begin{tabular}{|c|c|c|}
    \hline
        Hyperparameter &  Minimum Value & Maximum Value\\ \hline
        LTE tolerance ($\gamma$) & 0 & $10^{6}$\\
        \hline
    \end{tabular}
    \caption{Hyperparameter ranges for Adaptive FedECADO}
    \label{tab:hyperparameter_range_adaptive_fedecado}
\end{table}

\begin{table}[h!]
    \centering
    \begin{tabular}{|c|c|c|}
    \hline
        Hyperparameter &  Minimum Value & Maximum Value\\ \hline
        Global Step Size & 0 & $1$\\
        \hline       
        Local Step Size & 0 & $1$\\
        \hline
         Initial control variate ($c$) & 0 & $1$\\
        \hline
    \end{tabular}
    \caption{Hyperparameter ranges for SCAFFOLD}
    \label{tab:hyperparameter_range_adaptive_scaffold}
\end{table}

\begin{table}[h!]
    \centering
    \begin{tabular}{|c|c|c|}
    \hline
        Hyperparameter &  Minimum Value & Maximum Value\\ \hline
        Local Step Size & 0 & $1$\\
                \hline
    \end{tabular}
    \caption{Hyperparameter ranges for FedExp}
    \label{tab:hyperparameter_range_adaptive_fedexp}
\end{table}

\begin{table}[h!]
    \centering
    \begin{tabular}{|c|c|c|}
    \hline
        Hyperparameter &  Minimum Value & Maximum Value\\ \hline
        Global Step Size & 0 & $1$\\
        \hline   
        Local Step Size & 0 & $1$\\
                \hline
        Momentum Factor $\beta_1$ & 0.9 & $1$\\
                \hline       
        Momentum Factor $\beta_2$ & 0.9 & $1$\\
                \hline
    \end{tabular}
    \caption{Hyperparameter ranges for FedAdam}
    \label{tab:hyperparameter_range_adaptive_fedadam}
\end{table}

\begin{table}[h!]
    \centering
    \begin{tabular}{|c|c|c|}
    \hline
        Hyperparameter &  Minimum Value & Maximum Value\\ \hline
          Global Step Size & 0 & $1$\\
        \hline   
        Local Step Size & 0 & $1$\\
                \hline
        Momentum Factor $\beta$ & 0.9 & $1$\\
                        \hline
    \end{tabular}
    \caption{Hyperparameter ranges for FedAdaGrad}
    \label{tab:hyperparameter_range_adaptive_fedadagrad}
\end{table}

\section{Heterogeneous Federated Learning for Sentiment-140}
\label{app:sentiment_140}
\begin{figure}[h!]
    \centering
    \includegraphics[width=0.8\linewidth]{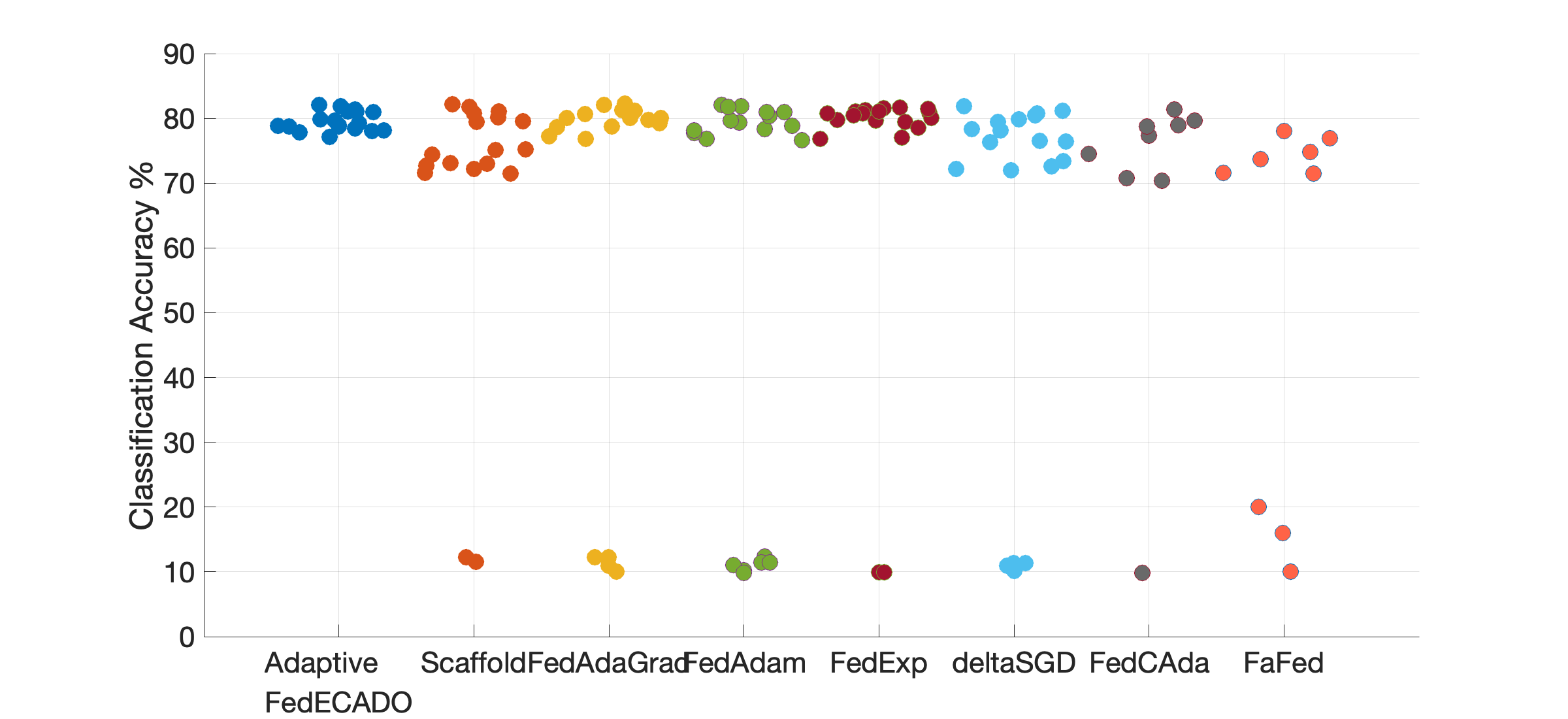}
    \caption{Hyperparameters for Adaptive FedECADO, SCAFFOLD, FedAdam, FedAdaGrad, FedExp and deltaSGD are swept during heterogeneous training for LSTM model for Sentiment140 dataset. The resulting classification accuracies of the random sweep are illustrated}
    \label{fig:sentiment_plot}
\end{figure}

We train a VGG-11 model on a Sentiment-140 dataset across 50 clients with an active client particition of 20\%. We vary the hyperparameters of each adaptive federated learning method according to Appendix \ref{app:hyperparameter_ranges} and denote the final classification accuracies in Figure \ref{fig:sentiment_plot}.

\begin{table}[h!]
    \centering
    \resizebox{1\textwidth}{!}{%
    \begin{tabular}{|p{18mm}|p{18mm}|p{18mm}|p{15mm}|p{18mm}|p{15mm}|p{15mm}|p{15mm}|p{15mm}|}
    \hline
    Dataset (Model) & Adaptive FedECADO & SCAFFOLD &FedAdam& FedAdaGrad & FedExp & deltaSGD & FedCAda & FaFed\\
    \hline
        CIFAR-10 \newline (ResNet-18) & 81.9 (2.1) & 54.5 (34.6) & 21.2 (25.8) & 20.4 (21.3) & 27.7 (12.6) & 44.7 (34.7) & 60.1 (29.2) & 62.9 (29.9) \\
        \hline
        CIFAR-100 (ResNet-50) & 50.4 (1.37) & 13.4 (12.3) & 6.5 (1.2) & 4.31 (8.21) & 3.6 (6.2) & 31.6 (10.9) & 33.3 (21.6)& 32.2 (19.7)\\ \hline
        Sentiment-140 (VGG-11) & 79.6 (1.49) & 73.32 (21.2) & 64.5 (29.3) & 62.4 (29.1) & 72.7 (22.9) & 63.4 (27.8) & 69.1 (22.5) & 54.7 (29.7) \\
        \hline
             \end{tabular}%
             }
        \caption{ Mean (Standard Deviation) of classification accuracies across all hyperparameter selections for training CIFAR-10, CIFAR-100, and Sentiment-140 datasets.}
    \label{tab:hyperparameter_sweep_percentages}

\end{table}
\section{Perturbing Hyperparameter Selections}
\label{app:perturbing_hyperparameters}
We study the effect of perturbing hyperparameters from their optimal values. The optimal hyperparemter selections for training CIFAR-10 on ResNet-18 model is determined from a hyperparameter sweep, with the final test accuracies of the sweep presented in Figure \ref{fig:test}. We then perturb the hyperparameter values by 20\% to see the effect on the model performance. As shown in Table \ref{tab:perturb_hyperparameters}, Adaptive FedECADO is highly insensitive to any perturbations where as many comparison methods can have large model performances. 

\begin{figure}[h!]
    \centering
    \includegraphics[width=0.4\linewidth]{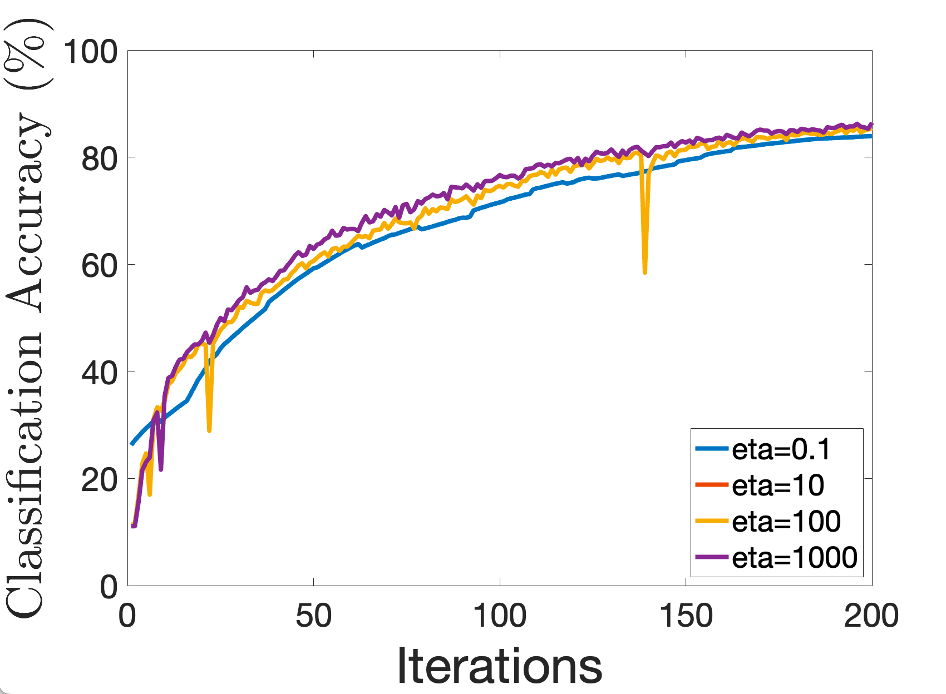}
    \caption{\small The hyperparameter of Adaptive FedECADO, $\gamma$ is swept across four order of magnitude in training CIFAR-10 dataset with ResNet-18 model under heterogeneous conditions. A wide range of $\gamma$ results in similar convergence plots.}
    \label{fig:eta_sweep}
\end{figure}

The insensitivity of Adaptive FedECADO's single hyperparameter ($\gamma$) is especially highlighted by varying the value of $\gamma$ by four orders of magnitude in Figure \ref{fig:eta_sweep}, where we observe stable and efficient global convergence. The insensitivity to $\gamma$ is partly due to using a numerically stable Backward-Euler integration method for aggregating client updates.

\begin{table}[h!]
    \centering
    \begin{tabular}{|p{18mm}|p{20mm}|c|c|c|c|c|c|c|}
    \hline
    CIFAR-10 (Resnet-18) & Adaptive FedECADO & SCAFFOLD & FedAdam & FedAdaGrad & FedExp & deltaSGD& FedCAda & FaFed\\
    \hline
        Mean (Std) & 84.5 (1.3) & 85.1 (3.2) & 75.1 (10.2) & 70.4 (8.8) & 80.4 (6.4) & 82.2 (5.7) & 82.3 (6.4)& 80.9 (4.9)\\
        \hline
             \end{tabular}

    \caption{Distribution of classification accuracies of federated learning methods trained under heterogeneous settings with hyperparameters randomly selected within 20\% of the optimal values.}
    \label{tab:perturb_hyperparameters}
\end{table}

\section{Effect of Client Schedulers}
\label{app:schedulers_experiment}

In non-federated settings, schedulers are often used to adjust learning rates to achieve stable convergence. However, the challenge in including schedulers for each client in federated settings is that this creates non-uniform step sizes that requires proper coordination with the central agent aggregation step. Additionally, schedulers introduce their own hyperparameters that influence convergence rate, creating larger space of potential hyperparameter selections.

\begin{figure}
    \centering
    \includegraphics[width=0.7\linewidth]{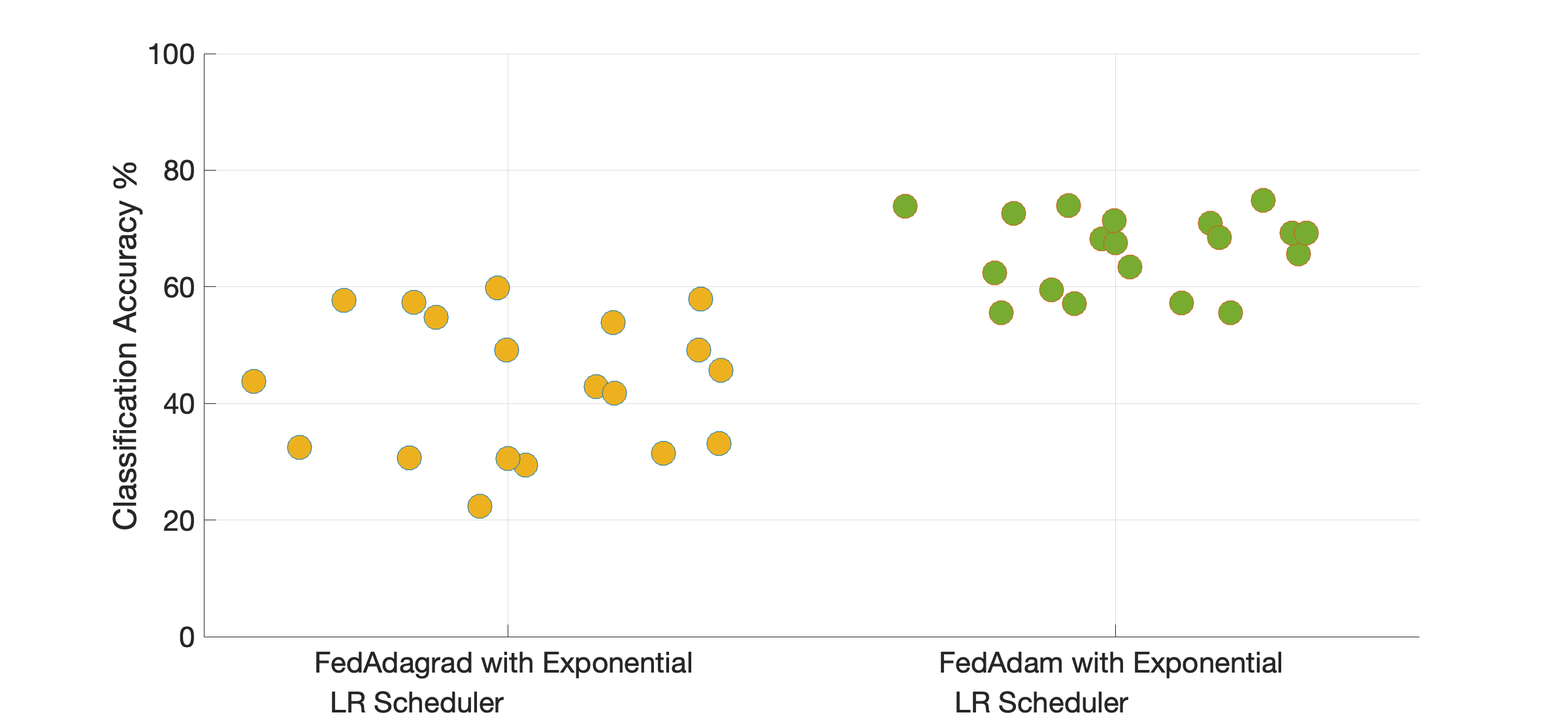}
    \caption{An exponential learning rate scheduler is added to the client updates with server aggregation performed by FedAdam and FedAdagrad \cite{reddi2020adaptive}. The hyperparameters of the server aggregation and client schedulers are randomly selected and the final classification accuracies of each run is shown.}
    \label{fig:scheduler_sweep}
\end{figure}

To demonstrate the challenge of introducing schedulers, we add exponential learning rate scheduler \cite{li2019exponential} to client updates of FedAdaGrad and FedAdam aggregations and perform a random search of hyperparameter selections. The results, shown in Figure \ref{fig:scheduler_sweep}, indicate that schedulers results in fewer divergent cases, but slow the overall convergence for achieving a high-performing model. The hyperparameter for the exponential learning rate scheduler ($\gamma$) is randomly sampled from a uniform distribution, $U$ as $\gamma \sim U[0,1]$.

\section{Approximating Hessian Computation}
\label{app:hessian}

Adaptive FedECADO requires computing the Hessian of each client via a smaller sample set before training to compute the optimal inductance parameters, $L_r$. However, computing the full Hessian may become computationally infeasible for large-scale models.
 In such circumstances, we can extend our framework to use Hessian approximations such as diagonal Fisher matrix \cite{jhunjhunwala2024fedfisher}. We note that the main goal of the Hessian is not to accurately characterize the loss function, but is instead used as a relative weighting between clients. As a result, approximate hessian methods can still be applied with relatively high accuracy.

\begin{figure}
     \centering
     \includegraphics[width=0.5\linewidth]{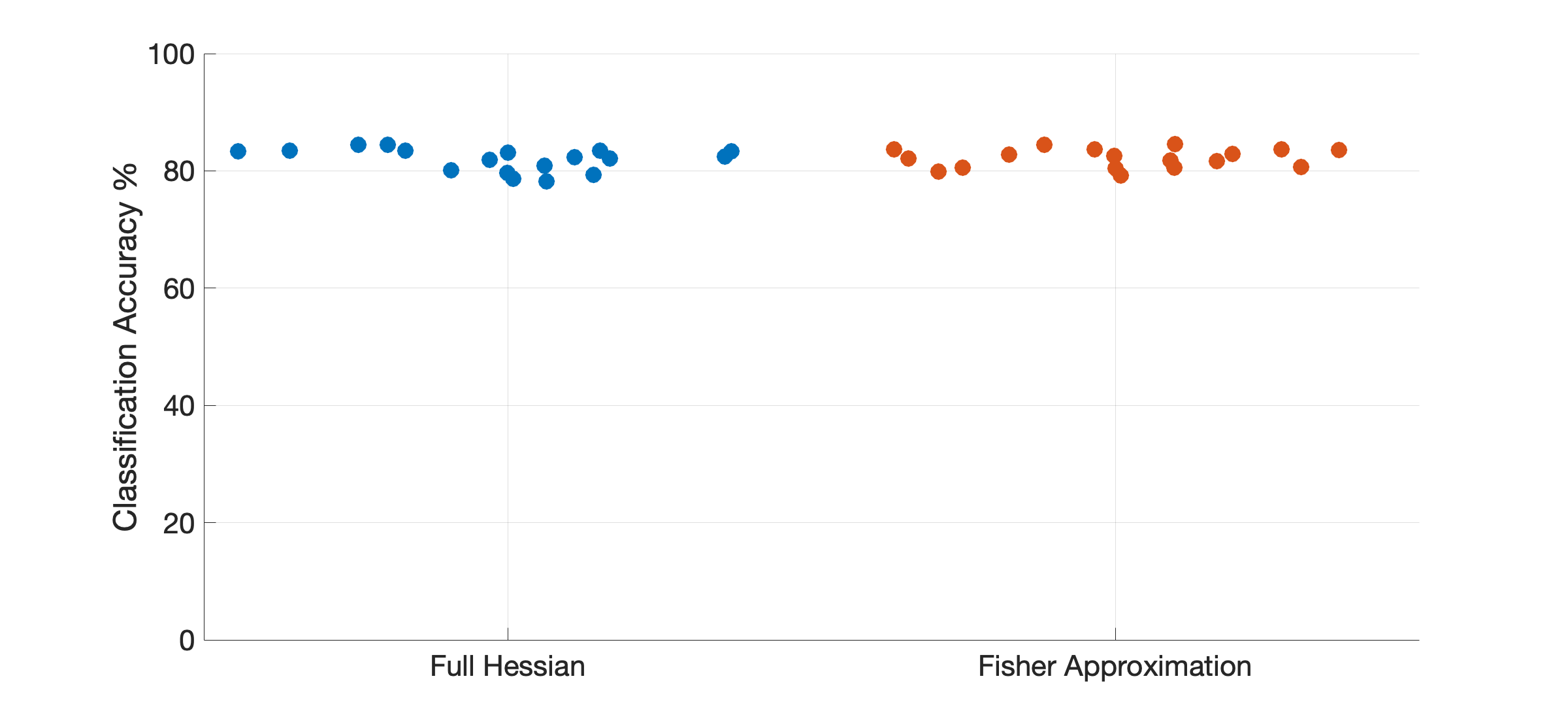}
     \caption{A ResNet-18 model is trained for CIFAR-10 dataset with Adaptive FedECADO using (a) full Hessian and (b) Fisher matrix approximation \cite{jhunjhunwala2024fedfisher} to compute the chord model.}
     \label{fig:hessian_comp}
 \end{figure}

 In this experiment, we study the the robustness of Adaptive FedECADO for training CIFAR-10 dataset on ResNet-18 model with random selections of LTE tolerance, $\eta$ using both full Hessian and approximate Hessian via Fischer matrix \cite{jhunjhunwala2024fedfisher}. Figure \ref{fig:hessian_comp} demonstrates that approximate Hessians offer similar performance to that of full Hessian. Our work provides a foundation for applying Hessian approximations in large models toward automated hyperparameter selection.

\section{Varying Configurations}
\label{app:varying_config}

To evaluate the generalizability of Adaptive FedECADO, we test its robustness under varying conditions of client participation and data heterogeneity. These experiments are designed to assess stability of our method under varying real-world settings. For each configuration, hyperparameters (listed in Appendix \ref{app:hyperparameter_ranges}) are randomly selected for both baseline methods and Adaptive FedECADO, and we report the mean and standard deviation of classification accuracy across 20 independent runs (Tables \ref{tab:resnet_variations_client_participation}–\ref{tab:resnet_variations_epochs}). The results demonstrate the robustness of Adaptive FedECADO, evidenced by its consistently low variance across runs, while also achieving competitive or superior mean classification accuracy compared to baseline methods.

\begin{table}[h!]
    \centering
    \begin{tabular}{|p{15mm}|p{20mm}|c|c|c|c|c|c|c|}
    \hline
  $[C,k]$ & Adaptive FedECADO & SCAFFOLD & FedAdam & FedAdaGrad & FedExp & deltaSGD& FedCAda & FaFed\\
    \hline
       $C=100$,$k=0.3$  & 84.5 (1.5) & 53.6 (34.6) & 26.8 (25.8) & 27.8 (21.3) & 39.1 (32.6) & 51.8 (34.7) & 35.0 (29.2) & 53.5 (29.9) \\
        \hline
       $C=100$,$k=0.1$  & 81.1 (2.3) & 57.2 (39.9) & 21.0 (35.3) & 42.9 (28.7) & 29.7 (44.6) & 41.0 (28.5) & 46.8 (17.4)& 57.2 (19.2)\\
        \hline
       $C=100$,$k=0.05$  & 78.1 (2.1) & 50.4 (34.1) & 20.6 (39.3) & 21.2 (40.2) & 33.7 (31.2) & 46.8 (19.5) & 32.0 (24.1)& 45.3 (27.9)\\
        \hline
       $C=20,k=0.3$  & 87.4 (3.5) & 56.7 (23.0) & 33.2 (32.2) & 22.8 (31.3) & 23.5 (46.4) & 57.4 (15.7) & 31.3  (23.4)& 57.2 (14.2)\\
        \hline
       $C=20,k=0.1$  & 84.7 (5.4) & 47.1 (33.2) & 30.3 (40.2) & 26.0 (28.8) & 24.8 (32.8) & 56.5 (22.1) & 39.6 (26.7)& 50.4 (15.1)\\
        \hline
             \end{tabular}

    \caption{  Mean (Std) Classification Accuracy of CIFAR-10 (Resnet-18) trained under varying client participation settings (number of clients, $C$, participation ratio, $k$) with hyperparameters of federated learning methods randomly selected within operational range.}
    \label{tab:resnet_variations_client_participation}
\end{table}

\begin{table}[h!]
    \centering
    \begin{tabular}{|p{18mm}|p{20mm}|c|c|c|c|c|c|c|}
    \hline
  $\delta$ & Adaptive FedECADO & SCAFFOLD & FedAdam & FedAdaGrad & FedExp & deltaSGD& FedCAda & FaFed\\
    \hline
        0.05 & 80.9 (1.3) & 53.6 (24.7) & 22.1 (36.1) & 29.2 (37.8) & 26.6 (38.6) & 53.5 (14.2) & 53.8 (19.4)& 57.4 (12.7)\\
        \hline
        0.1 & 84.2 (4.1) & 55.2 (23.2) & 25.0 (34.9) & 33.1 (26.6) & 26.5 (43.9) & 50.2 (15.8) & 60.6 (20.8)& 58.8 (14.2)\\
        \hline
        0.5 & 86.6 (2.6) & 58.0 (20.8) & 37.5 (43.1) & 45.1 (26.3) & 33.1 (18.5) & 57.6 (21.0) & 62.5 (26.5)& 59.3 (13.3)\\
        \hline

             \end{tabular}

    \caption{  Classification Accuracy of CIFAR-10 (Resnet-18) trained under varying data distributions (Dir16$(\delta)$) with hyperparameters of federated learning methods randomly selected within operational range.}
    \label{tab:resnet_variations_data_distribution}
\end{table}

\begin{table}[h!]
    \centering
    \begin{tabular}{|p{18mm}|p{20mm}|c|c|c|c|c|c|c|}
    \hline
  $\bar{e}$ & Adaptive FedECADO & SCAFFOLD & FedAdam & FedAdaGrad & FedExp & deltaSGD& FedCAda & FaFed\\
    \hline
        80 & 87.4 (1.9) & 51.1 (32.1) & 39.8 (26.7) & 40.5 (18.1) & 40.2 (20.4) & 51.3 (27.7) & 51.2 (29.1)& 50.0 (24.2)\\
        \hline
        30 & 83.1 (3.5) & 46.1 (36.8) & 24.9 (42.8) & 39.4 (43.6) & 35.4 (37.8) & 55.2 (13.5) & 40.9 (28.6)& 39.8 (26.1)\\
        \hline
             \end{tabular}

    \caption{  Mean (Std) Classification Accuracy of CIFAR-10 (Resnet-18) trained under varying distributions of number of local epochs ($e_i\sim U(0,\bar{e}]$) with hyperparameters of federated learning methods randomly selected within operational range.}
    \label{tab:resnet_variations_epochs}
\end{table}

\section{Ablation Experiments}
\label{app:ablation}
The primary goal of our work is to demonstrate an adaptive federated learning algorithm for heterogeneous settings that achieves competitive model performance without manual hyperparameter tuning. Our approach adapts both the continuous-time trajectory via the inductance parameters, $L_i$, of the underlying dynamical system, and the discretization behavior through the choice of local client and central agent time-steps. Together, these ensure stable simulation and reliable convergence to the steady state.

From an optimization perspective, the two key degrees of freedom are the effective learning rates (local and central agent), and the momentum coefficients of each client.

While the choice of $L_i$ accelerates convergence in continuous time, it does not affect stability of the underlying dynamics. To validate this, we conducted an ablation study where momentum parameters were held constant across clients.

\begin{table}[]
    \centering
    \begin{tabular}{|c|c|}
    \hline
       Inductance Value, $L_i$  & Final Classification Accuracy  \\\hline
       Optimal Client-specific $L_i$ from \eqref{eq:critical_L}  & 85.1  \\\hline
       0.1 & 81.6 \\ \hline
        1 & 84.7 \\ \hline
        10 & 80.2\\ \hline 
    \end{tabular}
    \caption{Final classification accuracies of training ResNet-18 on CIFAR-10 under varying inductance parameters}
    \label{tab:inductance_ablation}
\end{table}

As shown in Table \ref{tab:inductance_ablation}, the performance changes slightly depending on the momentum value, but overall stability and final performance remain unaffected. This confirms that our method is robust in the absence of momentum tuning and stability is inherently preserved by the numerical discretization method.

\section{Wallclock Runtime}
\label{app:runtime}
The relative wall-clock times for all methods are presented in Table \ref{tab:runtime}. While our method incurs slightly higher computational cost due to the dynamical system formulation and Backward-Euler steps, we believe that this overhead is compensated by the elimination of the hyperparameter tuning phase, which can be time-consuming and costly in federated environments.

\begin{table}[h!]
    \centering
    \begin{tabular}{|c|c|}
        \hline
        Method & Normalized Wallclock Time \\
        \hline
        Adaptive FedECADO & 1 \\
        \hline
        FedAdam	& 0.94 \\
        \hline
        FedAdaGrad & 0.93 \\
        \hline 
        SCAFFOLD & 0.96 \\
        \hline
        FedExp & 0.94 \\
        \hline 
        deltaSGD & 0.97 \\
        \hline 
        FedCAda & 0.98 \\
        \hline 
        FAFED & 0.95 \\
        \hline
    \end{tabular}
    \caption{Wallclock runtime (normalized to Adaptive FedECADO) for training a ResNet-18 model on the CIFAR-10 dataset over 100 epochs using different baseline methods.}
    \label{tab:runtime}
\end{table}

\end{document}